\documentclass{IEEEtran}

\usepackage[cmex10]{amsmath}
\usepackage{amssymb}
\usepackage{amsfonts}

\usepackage{amsthm}
\usepackage{algorithm}
\usepackage{algpseudocode}
\usepackage{graphicx}
\usepackage{epsfig}
\usepackage{cite}
\usepackage{tensor}
\usepackage[caption=false,font=footnotesize]{subfig}
\usepackage{color}

\usepackage{hyperref}
\hypersetup{hypertex=true,
	colorlinks=true,
	linkcolor=blue,
	anchorcolor=blue,
	citecolor=blue}

\usepackage{makecell}
\usepackage{float}
\usepackage{comment}
\usepackage{subfig}
\usepackage[dvipsnames]{xcolor}

\graphicspath{{figures/}}
\newtheorem{theorem}{Theorem}

\theoremstyle{definition}
\newtheorem{remark}{Remark}

\newcommand\T{{\hspace{-0pt}\intercal}}

\DeclareMathOperator{\diag}{diag}

\usepackage{ulem}
\usepackage{cancel}

\begin{document}
\normalem
\title{Adaptive Shape Servoing of Elastic Rods using Parameterized Regression Features and\\Auto-Tuning Motion Controls}

\author{
	Jiaming~Qi$^\dag$, 
	Guangtao~Ran$^\dag$,~\IEEEmembership{Member,~IEEE},
	Bohui~Wang,~\IEEEmembership{Senior Member,~IEEE},
	Jian Liu,~\IEEEmembership{Member,~IEEE}, 
        Wanyu Ma,
        Peng Zhou, ~\IEEEmembership{Member,~IEEE}, and
	David Navarro-Alarcon,~\IEEEmembership{Senior Member,~IEEE}%
	\thanks{$^\dag$These authors contributed equality to this work. This work is supported by the Research Grants Council (RGC) under grant number 14203917 (Corresponding author: Guangtao Ran and Bohui Wang).}%
	\thanks{Jiaming Qi and Guangtao Ran are with Department of Control Science and Engineering, Harbin Institute of Technology, Harbin 150001, China. (e-mail: {qijm\_hit@163.com}; {ranguangtao@hit.edu.cn}).}%
        \thanks{Bohui Wang is with the School of Cyber Science and Engineering, Xi’an Jiaotong University, Xi’an 710049, China. (e-mail: {{wang31aa@126.com}}).}
        \thanks{Jian Liu is with the School of Automation, Southeast University, Nanjing 210096, China. (e-mail: {{bkliujian@seu.edu.cn}}).}
	\thanks{Wanyu Ma, Peng Zhou, and David Navarro-Alarcon are with the Department of Mechanical Engineering, The Hong Kong Polytechnic University, Kowloon, Hong Kong. (e-mail:~{{dna@ieee.org}}).}
	}

\maketitle


\begin{abstract}
	The robotic manipulation of deformable linear objects has shown great potential in a wide range of real-world applications. However, it presents many challenges due to the objects' complex nonlinearity and high-dimensional configuration.
	In this paper, we propose a new shape servoing framework to automatically manipulate elastic rods through visual feedback.
	Our new method uses parameterized regression features to compute a compact (low-dimensional) feature vector that quantifies the object's shape, thus, enabling to establish an explicit shape servo-loop.
    To automatically deform the rod into a desired shape, the proposed adaptive controller iteratively estimates the differential transformation between the robot's motion and the relative shape changes; 
    This valuable capability allows to effectively manipulate objects with unknown mechanical models.
    An auto-tuning algorithm is introduced to adjust the robot's shaping motions in real-time based on optimal performance criteria.
    To validate the proposed framework, a detailed experimental study with vision-guided robotic manipulators is presented.	
\end{abstract}



\section{Introduction}\label{section1}
\IEEEPARstart{T}{he}
manipulation of deformable objects is currently an open (and hot!) research problem in robotics \cite{yu2022global} that has attracted many researchers due to its applicability in many scenarios, e.g. positioning fabrics \cite{huang2022task}, 
shaping soft materials \cite{cherubini2020model}, 
assembling compliant components \cite{zhang2023visual}, manipulating deformable linear objects (DLO) \cite{qin2023dual}, etc.
The feedback control of the object's non-rigid configuration is referred in the literature as \emph{shape servoing} \cite{Journals:Sanchez2018}, which arises a frontier problem that presents three main challenges: 
(i) The efficient (viz. compact) characterization of its deformable shape (which is typically represented with high-dimensional visual feedback vectors); 
(ii) The identification of its motion-deformation model (which depends on the, generally unknown, mechanical properties of the object); 
(iii) The adaptive modulation of the robot's motions during the active shaping task (as soft objects might be delicate and thus easy to damage).

For shape servoing tasks, feature extraction is necessary to accurately describe the object with a small number of feedback coordinates \cite{navarro2017fourier}.
Traditional features to represent deformations include: centroids, distance, angles, curvature, etc \cite{navarro2014visual}.
However, these geometrically constructed features are local, thus, they cannot describe the overall shape of an object.
The development of global features presents an advantage in this problem.
Image moments were used in \cite{1321161} to characterize object's contours, yet, with a limited real-time performance due to its complicated calculations.
In \cite{Hu20193}, principal component analysis was used to project raw Fast Point Feature Histograms into a new space with higher variance and a lower dimension.
A catenary-based feature descriptor was developed for tethered mobile robots in \cite{laranjeira2017catenary}, however, it is only applicable to specific shapes.
A Fourier-based approach was developed in \cite{navarro2017fourier} to characterize contours with low-dimensional features.
B$\acute{\text{e}}$zier curves and Non-Uniform Rational Basis Splines (NURBS) are an interesting option to describe complex contours, however, they have not been sufficiently studied in the literature as a way to establish an explicit shape servo-loop.
Other types of representations rely on machine learning, e.g. \cite{nair2017combining}, however, these methods require very large data sets to generalize to different situations (which is difficult to guarantee in many applications).
The design of a computationally efficient feature extraction algorithm is a key problem in shape servoing.

To execute shape servoing tasks, a controller requires a model that captures the relationship between robot's motions and the produced object's deformations \cite{9623343}.
In \cite{navarro2013model}, a least-squares regression method was used to calculate the deformation Jacobian matrix; Yet, it requires prior knowledge of the object model to properly define a regression structure.
There are other approaches that do not require prior knowledge of the structure, e.g., in \cite{alambeigi2018autonomous} a Broyden-like method was used to online estimate this deformation matrix. 
Although these type of methods are easy to implement, they are prone to enter local minima.
Kalman Filter (KF) is well-known to estimate unknown variables using a series of measurements in the presence of noise and uncertainty.
Based on this idea, the method in \cite{Qian2002Online} constructed a state-space model with the elements of Jacobian matrix and used KF to adaptively estimate its unknown values.
Since the manipulation of deformable objects involves nonlinear models with considerable uncertainties, KF is a good option to robustly estimate these unknown terms.

Deformable objects exhibit more complex behaviours than rigid objects, therefore, it is advantageous to control its transient behavior.
This can be achieved through performance adjustment methods, which have been used in motor speed regulation problems \cite{ibrahim2019optimal} to achieve accurate tracking while considering various performance criteria (e.g., overshoot, rising, settling time) \cite{kamal2014speed}.
This idea is also useful in the manipulation of soft objects to achieve various dynamic requirements (e.g., soft objects may support rapid deformations, while stiffer materials may need to be slowly manipulated).
Performance adjustment can also help to avoid damage, e.g., due to over-stretching and over-compression of an object \cite{huang2021non}.
However, most of existing shape servoing methods adopt classical constant gain controls \cite{Journals:Sanchez2018}, which limit the types of dynamic responses they can achieve.

This paper presents a new vision-based framework to automatically deform elastic rods into desired shapes.
To quantify the configuration of these DLOs, we propose a general representation method based on parametric regression features, which enables to characterize shapes with a compact feedback-like vector.
To the best of authors’ knowledge, this is the first time that a shape controller uses B$\acute{\text{e}}$zier/NURBS features to establish an \emph{explicit} shape servo-loop (both features were used in \cite{Journals:Wang2018}, but only to project a contour into an image plane, which largely differs from our approach).
Our proposed method presents an adaptive control strategy that: (i) iteratively estimates the unknown deformation model (based on an unscented Kalman filter approach), and (ii) online adjusts the controller's parameters (which modulate the transient response).

The rest of this paper is organized as follows:
Section \ref{section2} presents the preliminaries;
Section \ref{section3} contains the derivation of the proposed method, and the stability of the system is proved using the Lyapunov theory.
In Section \ref{section4}, an experimental study is conducted to validate the proposed framework. Finally, the conclusions are presented in  Section \ref{sect6}.




\begin{figure}[t]
	\centering
	\includegraphics[scale=0.33]{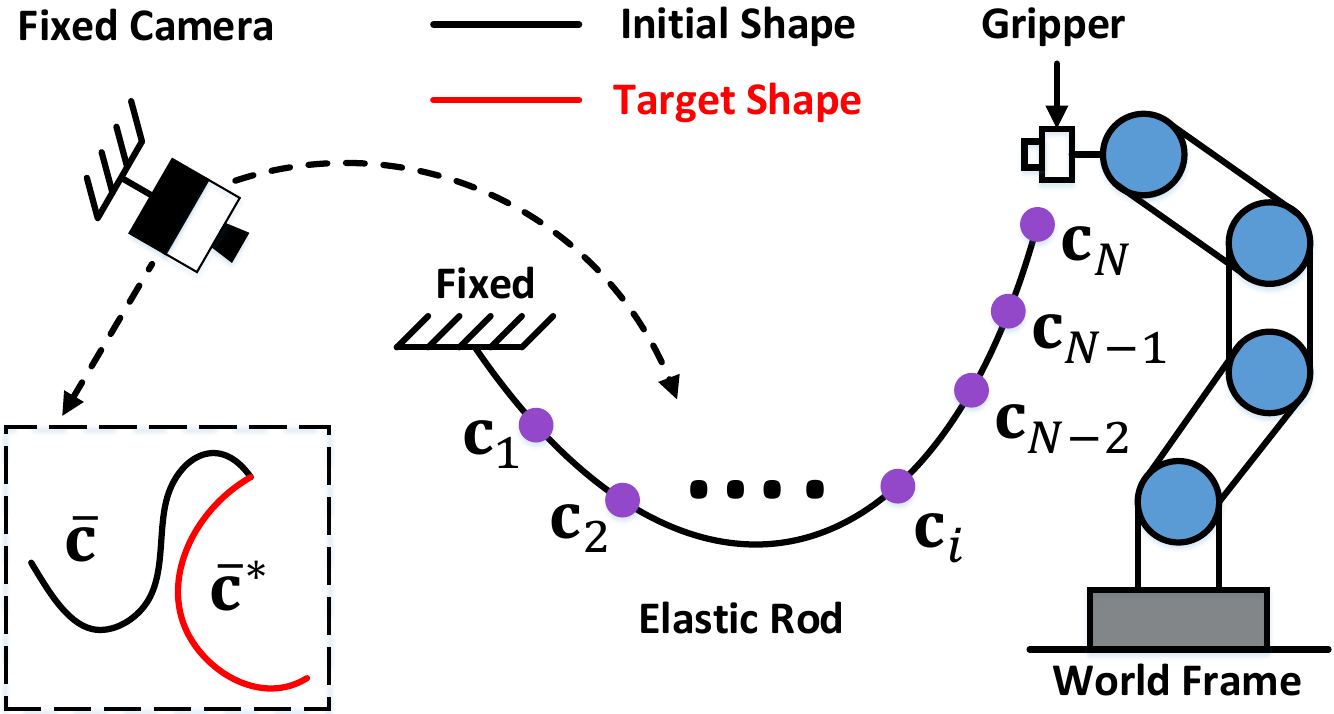}
	\vspace{-0.3cm}
	\caption{
 Representation of active manipulation of the elastic rod, where a vision sensor continuously measures the feature vector $\mathbf{s}$, which should be accurately deformed into target feature $\mathbf{s}^*$ within the controller.
	}
	\label{fig24}

 \vspace{-0.5cm}
\end{figure}

\vspace{-0.5cm}
\section{PRELIMINARIES}\label{section2}
\emph{Notation.}
Column vectors are denoted with bold small letters $\mathbf{m}$ and matrices with bold capital letters $\mathbf{M}$. 
Time evolving variables are represented as $\mathbf{x}_k$, where the subscript $k$ denotes the discrete time instant. 
$\mathbf{I}_n$ is an $n \times n$ identity matrix.
$\otimes$ represents the Kronecker product.

To derive our method, some conditions are now introduced:
\begin{itemize}
	\item 
	The 2D image feedback centerline of the rod is measured with a static camera in an eye-to-hand configuration (depicted in Fig. \ref{fig24}), and is denoted by:
	\begin{equation}
	\label{eq-1}
	{\bar{\mathbf{c}} = {{\left[ {\mathbf{c}_1^\T, \dots ,\mathbf{c}_N^\T} \right]}^\T} \in {\mathbb R^{2N}}}, \ {{\mathbf{c}_i} = {{\left[ {{c_{ui}},{c_{vi}}} \right]}^\T} \in {\mathbb R^2}}
	\end{equation}
	where $N$ is the number of the centerline points, $c_{ui}$ and $c_{vi}$ are the pixel coordinates of the $i$th image point.
	
	\item 
	During the manipulation task, the rod is rigidly grasped by the robot, and remains all the time within the observable range of the camera with no occlusions.
	
	\item The robot is commanded with classical kinematic controls $\Delta \mathbf{r}_k \in \mathbb{R}^q$ that render stiff behaviours and satisfy the incremental position $\mathbf{r}_{k} = \mathbf{r}_{k-1} + \Delta \mathbf{r}_k$, with bounded commands $\|\Delta\mathbf r_k\|\le \epsilon_r$, for $\epsilon_r>0$.
	
	\item The rod is manipulated slowly such that its shape is determined by the equilibrium of its potential/elastic energy terms only.
\end{itemize}

\textbf{Problem Statement}. 
Given a desired 2D constant centerline $\bar{\mathbf{c}}^*$, design a vision-based kinematic controller $\Delta \mathbf{r}_k$ to automatically deform an elastic rod such that its feedback shape $\bar{\mathbf{c}}$ approximates $\bar{\mathbf{c}}^*$, without identifying the mathematical model of the object or the camera's parameters.


\section{METHODS}\label{section3}

\subsection{Feedback Shape Parameters}
The \emph{na\"ive approach} to shape servoing is to synthesise a regulator that attempts to drive the full coordinates of $\bar{\mathbf c}$ into $\bar{\mathbf{c}}^*$.
The main problem within this approach is that $\bar{\mathbf{c}}$ is not efficient for real-time control as its dimension is very large.
Therefore, it is necessary to compute a reduced-dimension feature $\mathbf{s} \in \mathbb{R}^{p} (p \ll 2N)$, that can be used for feedback control.
Our proposed idea is to fit the 2D feedback centerline to a continuous curve $\mathbf{f}(\rho)\in\mathbb R^2$, for $\rho$ as a parametric variable representing the curve's normalized arc-length $ 0 \le \rho \le 1$.
Then, a point along the curve can be expressed as $\mathbf{c}_i=\mathbf{f}(\rho_i)$, with ${\rho}_i$ as the arc-length between the start point $\mathbf{c}_1$ and point $\mathbf{c}_i$ (see Fig. \ref{fig24}), for $\rho_1=0, \rho_N=1$.
The proposed parametric curve fitting is modelled as follows:
\begin{equation}
\label{eq2}
{\mathbf{f} \left( {{\rho}} \right) = \sum\limits_{j = 0}^n {{\mathbf{p} _j}{B_{j,n}}\left( {{\rho}} \right)}}
\end{equation}
where the positive scalar $n\in\mathbb N$ is the fitting order, and the vector $\mathbf{p}_j \in \mathbb{R}^2$ represents the shape parameters of $\bar{\mathbf{c}}$.
The function $B_{j,n}(\rho)$ models the chosen regression parameterization, which can take the following different forms:

\emph{1) Polynomial Parameterization.} It can easily represent smooth regular shapes, however, if $n$ is selected too large, curve overfitting may occur. The regression function is as follows:
\begin{equation}
\label{eq3}
{B_{j,n}}\left( {{\rho}} \right) = \rho^j
\end{equation}

\begin{figure}[t]
	\vspace{-0.7cm}	
	\centering
	\subfloat[B$\acute{\text{e}}$zier]
	{\includegraphics[scale=0.3]{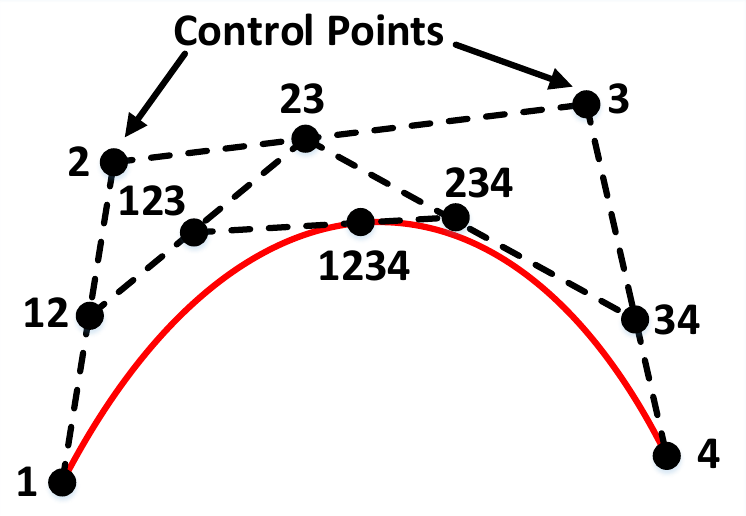}\label{fig24a}}
\hspace{0.3cm}	
 \subfloat[NURBS]
	{\includegraphics[scale=0.3]{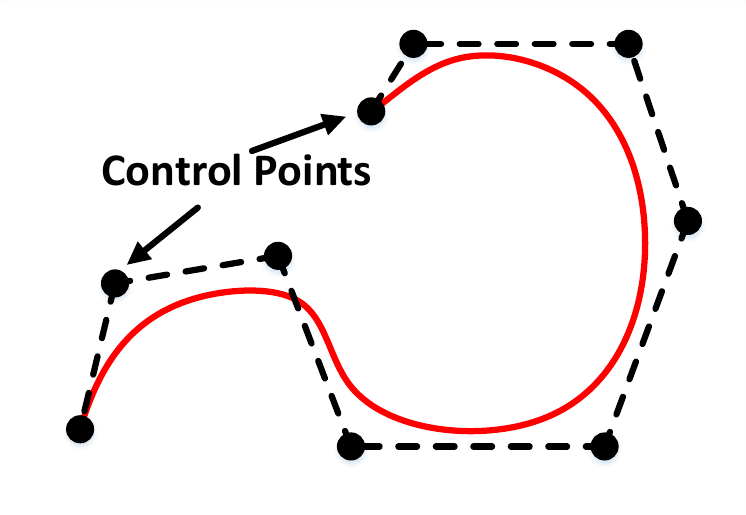}\label{fig24b}}

	\vspace{-0.21cm}
	\caption{
		Graphical representation of the B$\acute{\text{e}}$zier and NURBS curves.}
	\label{fig25}
	\vspace{-0.4cm}
\end{figure}

\emph{2) B$\acute{\text{e}}$zier Parameterization.}
As shown in Fig. \ref{fig25}, an $n$-degree B$\acute{\text{e}}$zier curve is defined with a polynomial expression using $n+1$ control points as follows:
\begin{equation}
\label{eq4}
B_{j,n}\left( {{\rho}} \right) = C_n^j{\left( {1 - {\rho}} \right)^{n - j}}\rho^j
\end{equation}
where $C_n^j$ denotes the binomial coefficients.
Here, $\mathbf{p}_j$ represents ``B\'ezier control points'', which are used to obtain a smooth fitting. 
When many control points are used, the degree of the curve and its computational cost are increased.

\emph{3) NURBS Parameterization \rm{\cite{piegl2012nurbs}}.}
It has local shape description properties, thus, the number of control points is independent from the degree of curves. 
This model can describe complex curves more accurately than polynomial/B\'ezier; Its regression function is as follows:
\begin{equation}
\label{eq5}
B_{j,n} \left( \rho \right) = \frac{N_{j,m} \left( \rho \right) \omega_j}{\sum_{l=0}^n N_{l,m}\left( \rho \right) \omega_l}   
\end{equation}
where $N_{j,m} (\rho)$ is the B\'ezier parameterization \eqref{eq4}, and ${\omega_j}$ are scalar weights.
Along this work, we set $m=n$ for simplicity.

\emph{4) Sinusoidal parameterization \rm{\cite{powell1981approximation}}.}
It is described as:
\begin{align}
\label{eq6}
{B_{j,n}}\left( {{\rho}} \right) = 
\left\{{
\begin{array}{*{20}{c}}
                           1,&{j = 0} \\
{\cos (\frac{j+1}{2} \rho)}, &{j > 0,\ j\ \rm{is}\ \rm{odd} } \\
{\sin (\frac{j}{2}   \rho)}, &{j > 0,\ j\ \rm{is}\ \rm{even}}
\end{array}} \right.
\end{align}
When $j$ is even, \eqref{eq6} denotes the well-known Fourier-based parameterization  \cite{navarro2017fourier}.



\begin{figure}[t]
	\centering
	\includegraphics[scale=0.21]{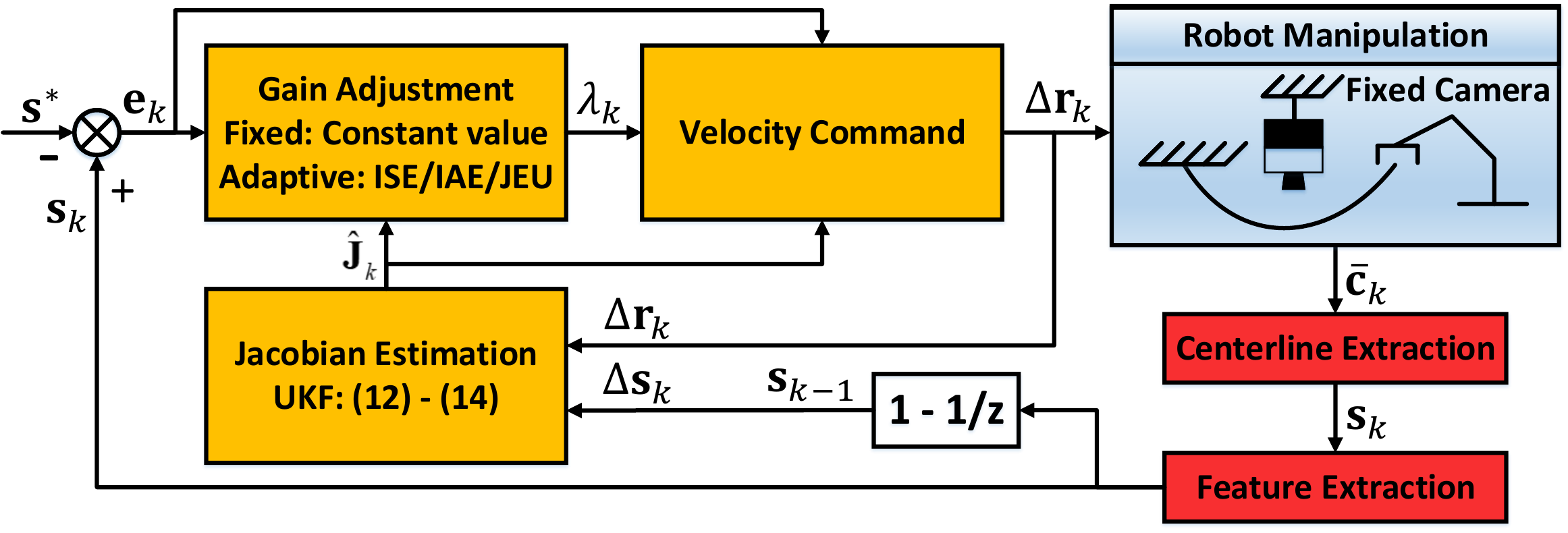}
	\vspace{-0.3cm}
	\caption{The block diagram showing the overall control workflow.}
	\label{fig23}
	\vspace{-0.47cm}
\end{figure}

By using the curve fitting model \eqref{eq2} (with any function in \eqref{eq3}--\eqref{eq6}), we can linearly parameterize the object's centerline as $\bar{\mathbf{c}} = \mathbf{B} \mathbf{s}$, for a ``tall'' regression matrix $\mathbf B$ satisfying:
\begin{align}
\label{eq65}
\mathbf{B} &= [\mathbf{B}_1^\T, \ldots, \mathbf{B}_N^\T]^\T 
\in \mathbb{R}^{2N \times 2(n+1)} \notag \\
\mathbf{B}_i &= [B_{0,n}(\rho_i),\ldots,B_{n,n}(\rho_i)] \otimes \mathbf{I}_2 \in \mathbb{R}^{2 \times 2(n+1)}  
\end{align}
and $\mathbf{s} = [\mathbf{p}_0^\T, \ldots, \mathbf{p}_n^\T]^\T  \in \mathbb{R}^{2(n+1)}$ as a vector of features that characterizes the shape.
This feature vector can be computed from sensor feedback at every iteration as follows:
\begin{equation}
\mathbf{s} = \boldsymbol {\mathcal{B}} \cdot \bar{\mathbf{c}},\ \  \text{for} \ \ \boldsymbol{\mathcal{B}} = {{\left( {{{{\mathbf{B}}}^\T}{\mathbf{B}}} \right)}^{ - 1}}{{{\mathbf{B}}}^\T}
\label{linear_parametrisation}
\end{equation}
To invert ${{\mathbf{B}}^\T}{{\mathbf{B}}}$, a sufficient number $N$ of data points must be used such that $2N > 2(n+1)$.
Fig. \ref{fig23} shows a schematic diagram of the overall manipulation strategy.


\begin{remark}
Although only four regression functions are given, there are other curve descriptors (e.g., B-spline, rational approximation, etc) that can be expressed as $\bar{\mathbf{c}} = \mathbf{B}\mathbf{s}$ to represent shapes with varying complexity.
\end{remark}

\subsection{UKF-Based Approximation of the Deformation Model}
Since we consider regular (i.e. mechanically well-behaved) elastic objects, it is reasonable to assume that small robot motions 
$\Delta \mathbf{r}_k$ will produce small changes $\Delta\bar{\mathbf{c}}_k$.
We locally model this situation (around the current operating point) with the expression, $\Delta\bar{\mathbf c}_k = \boldsymbol {\mathcal D}_k \cdot \Delta\mathbf r_k$
where $\boldsymbol {\mathcal D}_k$ is introduced to model the local deformation properties of the elastic object undergoing quasi-static manipulation by the robot.
By using this relation, we obtain the following motion model:
\begin{equation}
\label{eq-15}
\begin{array}{*{20}{c}}
\Delta{\mathbf s}_k = \boldsymbol {\mathcal B} \boldsymbol {\mathcal D}_k \cdot \Delta\mathbf r_k = \mathbf J_k \cdot  \Delta\mathbf r_k
\end{array}
\end{equation}
where $\Delta \mathbf{s}_k = \mathbf{s}_k - \mathbf{s}_{k-1} \in {\mathbb R^p}$ denotes the features' changes, and $\mathbf J_k = \boldsymbol {\mathcal B} \boldsymbol {\mathcal D}_k \in \mathbb R^{p \times q}$ represents a Jacobian-like matrix that transforms robot motions into feature changes;
This structure is difficult to analytically compute (i.e., with $\partial\mathbf s/\partial\mathbf r$), as the deformation properties of the DLO are generally \emph{unknown}.
Instead of identifying the full mechanical model, in this paper we design an algorithm that computes local approximations of $\mathbf J_k$ in real-time.
To this end, let us define the augmented state $\mathbf{x}_k = {\left[ {\partial {s_1}/\partial \mathbf{r}, \ldots ,\partial {s_p}/\partial \mathbf{r}} \right]^\T} \in {\mathbb R^{pq}}$, where $\partial {s_i} / \partial \mathbf r \in {\mathbb R^{1 \times q}}$ denotes the $i$th row of $\mathbf{J}_k$.
The discrete system \eqref{eq-15} can then be transformed into the state-space model:
\begin{equation}
\label{eq-17}
\begin{array}{*{20}{c}}
{\mathbf{x}_k = \mathbf{x}_{k-1} + \boldsymbol{\eta}_{k}},&
{\mathbf{y}_k = \mathbf{M}_k\mathbf{x}_k + \boldsymbol{\nu}_k}
\end{array}
\end{equation}
where the system output is defined as $\mathbf{y}_k = \Delta\mathbf{s}_k$, and which assumes the process and measurement noises to be zero-mean Gaussian white noise, i.e.,  $\boldsymbol{\eta}_k \sim N\left( {\mathbf{0},\mathbf{U}_k} \right)$ and  $\boldsymbol{\nu}_k \sim N\left( {\mathbf{0},\mathbf{W}_k} \right)$.
The measurement matrix $\mathbf{M}_k$ is defined as:
\begin{equation}
\label{eq-18}
\mathbf{M}_k = \diag \left( \Delta \mathbf{r}_k^\T,\cdots, \Delta \mathbf{r}_k^\T \right) \in {\mathbb R^{p \times pq}}
\end{equation}

We use a modified UKF \cite{xiong2006performance} (which has good dynamic performance and robustness to external disturbances) to compute local approximations of $\mathbf x_k$ in real-time.
UKF adopts the unscented transform (UT) to propagate mean and covariance through a nonlinear transformation among a minimal set of sample points (i.e., sigma points) \cite{liu2023image}. 
Then, the weighted sum method is used for the transformed samples to obtain the posterior mean and covariance of state vectors with a second-order accuracy.
The procedure can be summarized as follows:




Step 1: 
The variable $\mathbf{x}_{k-1}$ with mean $\widehat{\mathbf{x}}_{k-1}$ and covariance $\widehat{\mathbf{P}}_{k-1}$ is estimated by sigma points defined by
\begin{align}
\label{eq31}
{\chi}_{k-1}^0 &= \widehat{\mathbf{x}}_{k-1}, \ \
{\chi}_{k-1}^i = 
\widehat{\mathbf{x}}_{k-1} + ({a\sqrt {pq {\widehat{\mathbf{P}}_{k-1}}}})_i \notag\\
{\chi}_{k-1}^{i+pq} &= 
\widehat{\mathbf{x}}_{k-1} - ({a\sqrt {pq {\widehat{\mathbf{P}}_{k-1}}}})_i,
\ \  i=1,\ldots,pq
\end{align}
where ${\left( \bullet \right)_i}$ is the $i$th column of the cholesky matrix decomposition.
$a>0$ is a scaling parameter specifying the distribution of the sigma points around $\hat{\mathbf{x}}_{k-1}$.


Step 2: Prediction.
Sigma points pass through the process model to generate transformed points:
\begin{align}
\label{eq19}
\chi_{k|k-1}^i &= \chi_{k-1}^i,\  i=0,1,\ldots,2pq , \notag\\ \widehat{\mathbf{x}}_{k|k-1} &= \sum\limits_{i = 0}^{2pq}{\varpi_i{\chi}_{k|k-1}^i}  \\
\widehat{\mathbf{P}}_{k|k-1} &= 
\sum\limits_{i = 0}^{2pq} {\varpi_i 
	(\tilde{\chi}_{k|k-1}^i)
	(\tilde{\chi}_{k|k-1}^i)^\T} + ({\mathbf{U}_k} + {\Delta \mathbf{U}_k}) \notag 
\end{align}
where $\Delta \mathbf{U}_k>0$ is a positive-definite matrix to improve the approximation stability \cite{xiong2006performance}.
$\varpi_0 =  1- \frac{1}{a^2}, \varpi_i = \frac{1}{2pqa^2},i=1,\ldots,2pq$, are scalar weights and $\sum_{i=0}^{2pq}{\varpi_i}=1$. 
$\tilde{\chi}_{k|k-1}^i = {\chi}_{k|k-1}^i - \widehat{\mathbf{x}}_{k|k-1}$.

Step 3: Update.
Classical KF can be used to update the measurement signal as follows:
\begin{align}
\label{eq21}
\widehat{\mathbf{y}}_k &= \mathbf{M}_k \widehat{\mathbf{x}}_{k|k-1} \notag\\ 
\widehat{\mathbf{P}}_{yy,k} &= \mathbf{M}_k \widehat{\mathbf{P}}_{k|k-1} \mathbf{M}_k^{\T} + \mathbf{W}_k \notag \\
\widehat{\mathbf{P}}_{{xy,k}} &= \widehat{\mathbf{P}}_{k|k-1} \mathbf{M}_k^{\T} \\
\widehat{\mathbf{x}}_{k} &= \widehat{\mathbf{x}}_{k|k-1} + \mathbf{K}_k \tilde{\mathbf{y}}_k  \notag \\
\widehat{\mathbf{P}}_{k} &= \widehat{\mathbf{P}}_{k|k-1} - \mathbf{K}_k{\widehat{\mathbf{P}}_{yy,k}}{\mathbf{K}_k^{\T}} \notag
\end{align}
where $\tilde{\mathbf{y}}_k = \mathbf{y}_k - \widehat{\mathbf{y}}_k$, and $\mathbf{K}_k = \widehat{\mathbf{P}}_{{xy,k}} \widehat{{\mathbf{P}}}_{yy,k}^{-1}$ is Kalman gain.


Step 4: Repeat steps 1--3 for the next update.

After completing an iteration, we de-vectorize $\widehat{\mathbf{x}}_k$ to get an estimation ${{\widehat{\mathbf{J}}}_k}$ of the unknown Jacobian matrix.


\begin{remark}
\label{remark2}
In practice, the robot's motion $\Delta \mathbf{r}_k$ is bounded, which implies that the elements of $\mathbf{M}_k$ are also bounded.
Also, note that the occlusion-free observations of the deformable object (with intrinsic regularity) imply that the noise matrices $\mathbf{U}_k$ and $\mathbf{W}_k$ are similarly bounded.
These conditions enable us to apply Theorem 1 from \cite{xiong2006performance} (we refer the reader to the original source), which ensures that (after the UKF's algorithm has been sufficiently iterated) the estimation error $\tilde{\mathbf{J}}_k =  \mathbf{J}_k - \widehat{\mathbf{J}}_k$ is bounded by a small positive scalar $\| \tilde{\mathbf{J}}_k \| \le \epsilon_J$.
This initialization of UKF can be easily achieved by performing small babbling motions at the starting configuration of the system, i.e., before the shape control experiment, see \cite{qi2021contour}.
\end{remark}

\begin{remark}
\label{remark3}
Throughout this paper, we assume that $\mathbf J_k$ and its estimation $\widehat{\mathbf{J}}_k$ are both full column rank.
This condition physically implies that the manipulated object is away from ill-conditioned configurations, e.g., around a fully-stretched shape or when the object suddenly breaks.
\end{remark}

\vspace{-0.3cm}
\subsection{Adaptive Shape Servoing Controller}
In this section, we design the motion command $\Delta \mathbf{r}_k$ to minimize the shape error ${\mathbf{e}_k = \mathbf{s}_k - \mathbf{s}^* }$ between the feedback feature $\mathbf{s}_k$ and a constant target $\mathbf{s}^*$.
To this end, let us introduce the following quadratic performance index:
\begin{equation}
\label{eq-28}
Q = \mathbf{e}_k^\T \mathbf{e}_k + \lambda \Delta \mathbf{r}_k^\T \Delta \mathbf{r}_k
\end{equation}
where ${\lambda}>0$ is a weight that can be used to specify the magnitude of the motion input ${\Delta \mathbf{r}_k}$.
From \eqref{eq-15}, we can obtain the following differential error model:
\begin{align}
\label{eq-26}
{\mathbf{e}_k} - {\mathbf{e}_{k - 1}} = \mathbf{J}_k \Delta {\mathbf{r}_k}, \ 
~{\mathbf{e}_k} + {\mathbf{e}_{k - 1}} = 2{\mathbf{e}_{k - 1}} + {\mathbf{J}_k}\Delta {\mathbf{r}_k}
\end{align}
which we use for deriving the control input.
To this end, let us compute the extremum $\partial Q/\partial\Delta \mathbf{r}_k = \mathbf 0$, replace $\mathbf{J}_k$ with $\widehat{\mathbf{J}}_k$, and solve the expression for $\Delta \mathbf{r}_k$: 
\begin{equation}
\label{eq30}
\Delta {\mathbf{r}_k} = -  \boldsymbol\Phi ^{ - 1} \widehat{\mathbf{J}}_k^{\T}{\mathbf{e}_{k - 1}}, \quad
{\boldsymbol\Phi } = {\lambda }{\mathbf{I}_q} + \widehat{\mathbf{J}} _k^{\T}{{\widehat{\mathbf{J}} }_k}
\end{equation}
The performance of this controller can be regulated by adjusting the values of the positive gain $\lambda$.
This valuable property is beneficial when manipulating soft/delicate objects during the shaping motions.
For example, for $\lambda \approx 0$, the control input takes the form $\Delta \mathbf{r}_k \approx - \widehat{\mathbf{J}}_k^{+}\mathbf{e}_{k-1}$ with the largest magnitude, where $\widehat{\mathbf{J}}_k^{+}$ denotes the standard pseudo-inverse matrix.
While $\lambda \to \infty$, $\Delta \mathbf{r}_k \approx \mathbf{0}$ makes the system converge slowly.
Intuitively speaking, manually setting $\lambda$ can flexibly make the system adapt to various working conditions.
However, simply increasing $\lambda$ may cause the slow motion, while decreasing $\lambda$ may cause overshoot (due to the excessive error).
Therefore, $\lambda$ should be adjusted automatically to always be optimal to achieve the ideal dynamic response.
For this issue, a parameter optimization algorithm based on the gradient-descent method to adjust $\lambda$ is given as follows:

\emph{i) ISE (integral squared error):}
It is a well-known criteria for obtaining optimal controller parameters \cite{Zhuang1993Automatic}. It penalizes large errors with the metric:
\begin{equation}
\label{eq-39}
H =\frac{1}{2} \sum\limits_{k = 1}^{\infty} {\mathbf{e}_k^\T{\mathbf{e}_k}}
\end{equation}
which is used for computing the parameter tuning criterion:
\begin{align}
\label{eq-40}
\nabla_{\lambda}H 
&= \sum\limits_{k = 1}^{\infty}
{\mathbf{e}_{k - 1}^\T{{\widehat {\mathbf{J}}}_k}\left( {{\mathbf{I}_q} - {\boldsymbol\Phi ^{ - 1}}\widehat {\mathbf{J}}_k^\T{{\widehat {\mathbf{J}}}_k}} \right){\boldsymbol\Phi ^{ - 2}}\widehat {\mathbf{J}}_k^\T{\mathbf{e}_{k - 1}}}
\end{align}

\emph{ii) IAE (integral absolute error):}
It defines a metric based on the \emph{magnitude} of the error
\begin{equation}
\label{eq-42}
H = \sum\limits_{k = 1}^{\infty} {\left\| {{\mathbf{e}_k}} \right\|}
\end{equation}
which makes it sensitive to small errors. 
The parameter tuning criterion is defined as follows:
\begin{equation}
\label{eq-43}
\nabla_{\lambda}H 
= \sum\limits_{k = 1}^{\infty} 
{\frac{{\mathbf{e}_{k - 1}^{\T}{{\widehat {\mathbf{J}}}_k}\left( {{\mathbf{I}_q} - {\boldsymbol\Phi ^{ - 1}}\widehat {\mathbf{J}}_k^{\T}{{\widehat {\mathbf{J}}}_k}} \right){\boldsymbol\Phi ^{ - 2}}\widehat {\mathbf{J}}_k^{\T}{\mathbf{e}_{k - 1}}}}{\|\mathbf{e}_k\|}} 
\end{equation}

\emph{iii) JEU (integral of the squared error and control)}
ISE and IAE only consider error-based performance constraints, thus, cannot adjust other system requirements, e.g., minimum overshoot, peak time, rising time.
The JEU criterion contains a weighted sum of both squared error and control terms \cite{Xiong2005Study}:
\begin{equation}
\label{eq-44}
H = \frac{1}{2}\sum\limits_{k = 1}^{\infty} \left( {{\omega _1}\mathbf{e} _k^\T{\mathbf{e}_k} + {\omega _2}\Delta \mathbf{r}_k^\T\Delta {\mathbf{r}_k}} \right)
\end{equation}	
for ${\omega}_1$ and ${\omega}_2$ as positive scalars that satisfy ${\omega}_1+{\omega}_2=1$. 
The JEU-based parameter tuning criterion is: 
\begin{equation}
\label{eq35}
\nabla_{\lambda}H 
= \sum\limits_{k = 1}^{\infty} {\left( {{\omega _1}\mathbf{e}_k^\T{{\widehat{\mathbf{J}}}_k} + {\omega _2}\Delta \mathbf{r}_k^\T} \right){\boldsymbol\Phi^{-2}}\widehat{\mathbf{J}}_k^\T\mathbf{e}_{k-1}}
\end{equation}

\begin{remark}
Note that for the above three criteria, we replace the unknown deformation Jacobian matrix $\mathbf{J}_k$ by its numerical approximation $\widehat{\mathbf{J}}_k$. 
\end{remark}
In our proposed method, the control parameter ${\lambda}$ is dynamically updated with the gradient-descent rule as follows:
\begin{equation}
\label{eq33}
{{{\lambda} _k} = {{\lambda} _{k-1}} - d \cdot \nabla_{\lambda}H},\ \ \text{for} \ \lambda_k \ge \epsilon_\lambda
\end{equation}
where $d$ is a positive gain to control the update rate of $\lambda_k$, and $\epsilon_\lambda>0$ is a small scalar.
The initial value $\lambda_0$ is generally set to a large positive value to prevent control saturation (arising e.g., due to a large initial error $\mathbf e_k$).
These tuning criteria are only implemented when $\lambda_k \ge \epsilon_\lambda$ to prevent it becoming smaller than $\epsilon_\lambda$.
This adaptive update rule is used to compute $\lambda$ in the control law \eqref{eq30}.

\begin{theorem}
Consider the dynamic system \eqref{eq-15} in closed-loop with the controller \eqref{eq30}, with model estimator \eqref{eq31}--\eqref{eq21} and gain adjustment \eqref{eq33}.
For this system, the deformation error $\mathbf{e}_k$ converges to a compact set around zero, which is uniformly bounded.
\end{theorem}



\begin{IEEEproof}
Consider the discrete Lyapunov function defined by $V_k = \frac{1}{2}\mathbf{e}_k^\T\mathbf{e}_k$, whose finite difference is \cite{Sarpturk1987On}:
\begin{align}
\label{eq-35}
\Delta V_k &= V_k - V_{k - 1} = \frac{1}{2}\mathbf{e}_k^\T\mathbf{e}_k - \frac{1}{2}\mathbf{e}_{k - 1}^\T\mathbf{e}_{k - 1}
\end{align}
By substituting \eqref{eq-26} and \eqref{eq30} into \eqref{eq-35}, we obtain:
\begin{align}
\label{eq-36}  
\Delta {V_k}
&= ( {{\mathbf{e}_{k - 1}} + \tfrac{1}{2}{{{\mathbf{J}}}_k}\Delta {\mathbf{r}_k}} )^\T{{{\mathbf{J}}}_k}\Delta {\mathbf{r}_k} \notag \\
&= \mathbf{e}_{k - 1}^\T {\mathbf{J}_k}\Delta {\mathbf{r}_k} + \frac{1}{2}\Delta \mathbf{r}_k^\T \mathbf{J}_k^\T {\mathbf{J}_k}\Delta {\mathbf{r}_k} \\
&= \mathbf{e}_{k - 1}^\T {{\widehat{\mathbf{J}} }_k}\Delta {\mathbf{r}_k} + \mathbf{e}_{k - 1}^\T{{\tilde{\mathbf{J}} }_k}\Delta {\mathbf{r}_k} + \frac{1}{2}\Delta \mathbf{r}_k^\T \mathbf{J}_k^\T{\mathbf{J}_k}\Delta {\mathbf{r}_k} \notag \\
&=  - \mathbf{e}_{k - 1}^\T{\mathbf{L}_k}{\mathbf{e}_{k - 1}} + \mathbf{e}_{k - 1}^\T{{\tilde{\mathbf{J}} }_k}\Delta {\mathbf{r}_k} + \frac{1}{2}\Delta \mathbf{r}_k^\T \mathbf{J}_k^\T{\mathbf{J}_k}\Delta {\mathbf{r}_k} \notag
\end{align}
for $\mathbf{L}_k={{\widehat{\mathbf{J}} }_k}{\boldsymbol \Phi ^{ - 1}}\widehat{\mathbf{J}} _k^\T\ge 0$ as a positive semi-definite symmetric matrix.
With the aim to establish the convergence set $\Omega$ where $\Delta V_k \le 0$, let us analyse the following relation:
\begin{equation}
    \label{eq32}
    - \mathbf{e}_{k - 1}^\T{\mathbf{L}_k}{\mathbf{e}_{k - 1}} + \mathbf{e}_{k - 1}^\T{{\tilde{\mathbf{J}} }_k}\Delta {\mathbf{r}_k} + \frac{1}{2}\Delta \mathbf{r}_k^\T \mathbf{J}_k^\T{\mathbf{J}_k}\Delta {\mathbf{r}_k} \le 0
\end{equation}
Considering the Young's inequality and the upper bounds of $\tilde{\mathbf{J}}_k$ and $\Delta \mathbf{r}_k$ (see Sec. \ref{section2} and Remark \ref{remark2}), we can obtain:
\begin{equation}
    \label{eq36}
    - \mathbf{e}_{k - 1}^\T{{\tilde{\mathbf{J}} }_k}\Delta {\mathbf{r}_k} \le \frac{1}{2}{\left\| {{\mathbf{e}_{k - 1}}} \right\|^2} + \frac{1}{2}\epsilon _J^2\epsilon _r^2
\end{equation}
Substituting \eqref{eq36} into \eqref{eq32}, and after some algebraic and bounds operations, the following relations can be derived:
\begin{align}
    \label{eq37}
    \Delta \mathbf{r}_k^\T \mathbf{J}_k^\T {\mathbf{J}_k}\Delta {\mathbf{r}_k} &\le \left( {2{\bar{\beta} ({{\mathbf{L}_k}}) } + 1} \right){\left\| {{\mathbf{e}_{k - 1}}} \right\|^2} + \epsilon _J^2\epsilon _r^2 \\
    \label{eq38}
    \epsilon _r^2\bar \beta \left( {\mathbf{J}_k^\T {\mathbf{J}_k}} \right) - \epsilon _J^2\epsilon _r^2 &\le \left( {2\bar \beta \left( {{\mathbf{L}_k}} \right) + 1} \right){\left\| {{\mathbf{e}_{k - 1}}} \right\|^2}
\end{align}
where $\bar{\beta}(\bullet)$ is the maximum eigenvalue of $\bullet$.
Finally, the convergence set $\Omega$ of $\mathbf{e}_k$ is calculated as follows:
\begin{equation}
    \label{eq39}
    \Omega = \left\{ \mathbf e_k ~{\big|}~2V_k=
    {\left\| {{\mathbf{e}_{k}}} \right\|^2} \le \frac{{\epsilon _r^2\left( {\bar \beta \left( {\mathbf{J}_k^\T {\mathbf{J}_k}} \right) - \epsilon _J^2} \right)}}{{2\bar \beta \left( {{\mathbf{L}_k}} \right) + 1}} \right\}
\end{equation}
It is assumed that UKF has been sufficiently iterated to provide a small enough $\epsilon_J$ that ensures that ${\bar \beta \left( {\mathbf{J}_k^\T {\mathbf{J}_k}} \right) - \epsilon _J^2} > 0$ (see Remark \ref{remark2}).
From \eqref{eq39}, we can see that $\mathbf{e}_k$ is uniformly bounded \cite{ma2014adaptive} with the convergence set $\Omega$.
As $\lambda_k$ remains positive semi-definite, $\mathbf{L}_k$ will always remain positive semi-definite, this ensures that $\bar{\beta}(\mathbf{L}_k)$ is always positive.
Thus, the stability of the system is not affected (considering that both the numerator and denominator are greater than zero), i.e., $\|\mathbf{e}_k\|$ will converge into $\Omega$.

\end{IEEEproof}

\vspace{-0.2cm}
\begin{remark}
For ISE \eqref{eq-40}, $\lambda_k$ decreases (as $\nabla_{\lambda}H \ge 0$) to make the system converge faster with enlarging $\Delta \mathbf{r}_k$.
For IAE \eqref{eq-43}, $\lambda_k$ is also reduced due to positive $\nabla_{\lambda}H$.
When $\|\mathbf{e}_k\|$ is large, IAE is slower than ISE, but when $\|\mathbf{e}_k\|$ is small (especially for $\|\mathbf{e}_k\|\le 1$), the acceleration effect of IAE will be more obvious than that of ISE to update $\lambda_k$ faster.
For JEU \eqref{eq35}, it is similar to ISE when $\omega_1=1$ and $\omega_2=0$.
While, for the case of $\omega_1=0, \omega_2=1$, $\lambda_k$ increases (as $\nabla_{\lambda}H \le 0$) to limit $\Delta \mathbf{r}_k$, which can be explained that the elastic rod is deformed in a slower manner.
When neither is zero, the elastic rod is deformed with the given weights considered.
\end{remark}


\vspace{-0.2cm}
\begin{remark}
The proposed framework cannot guarantee to deform all target shapes.
As $\mathbf{L}_k$ is never full rank, for unfeasible targets, the feedback shape can only converge to nearby neighborhoods of $\mathbf s^*$.
For these types of overdetermined visual servoing controllers (i.e., with more output features than input controls), global asymptotic convergence of $\| \mathbf{e}_k \|$ cannot be guaranteed \cite{Journals:Chaumette2006}. 
\end{remark}

\begin{figure}[htbp]
    \centering
    \includegraphics[width=\columnwidth]{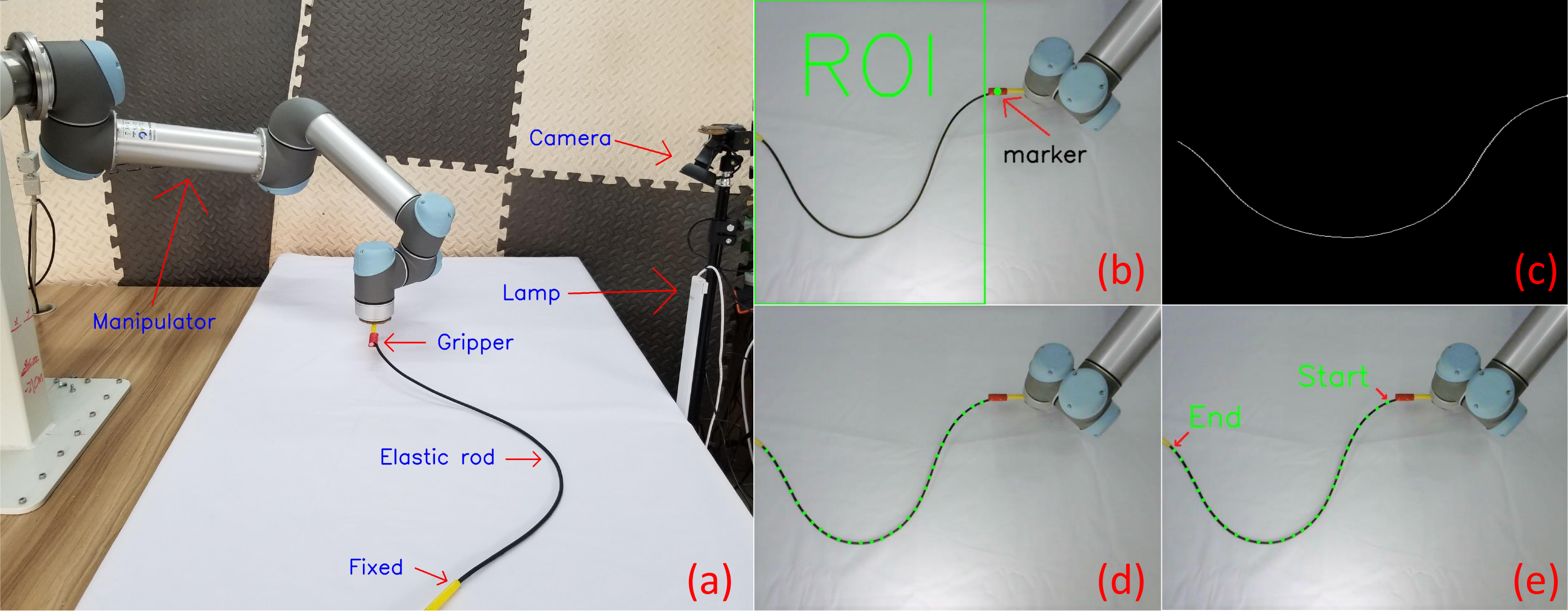}
    \vspace{-0.7cm}
    \caption{Experimental setup and image processing steps.}
    \label{fig11}
\end{figure}

\vspace{-0.4cm}
\section{Results}\label{section4}

\vspace{-0.3cm}
\subsection{Setup}\label{section4a}
This section validates the proposed regression features and the adaptive servoing controller.
As shown in Fig. 4a, the experimental platform includes an elastic rod (with one fixed end), a UR5 robot constrained to move over the horizontal ``xy'' plane,
and a static Logitech C270 camera to observe the scene.
Visual feedback is processed with the OpenCV libraries in a Linux-based PC at 30 fps.
A video with the conducted experiments can be downloaded from \url{https://github.com/q546163199/experiment_video/raw/master/paper1/video.mp4}.

The centroid of the gripper's red fiducial marker is highlighted by a green point, and is used to segment the region of interest (ROI) that contains the elastic rod (see Fig. 4b).
The \emph{OpenCV/Open} function used to denoise the ROI and obtain a binary image, which is further processed by the \emph{OpenCV/Thinning} function to obtain the unordered centerline of the rod's skeleton (see Fig. 4c).
Farthest Point Sampling is used to extract a fixed number of centerline points (see Fig. 4d).
The closest point to the gripper's marker is chosen as the starting point of the centerline parametric curve (see Fig. 4e).


\begin{figure}[htbp]
	\centering
	\includegraphics[scale=0.24]{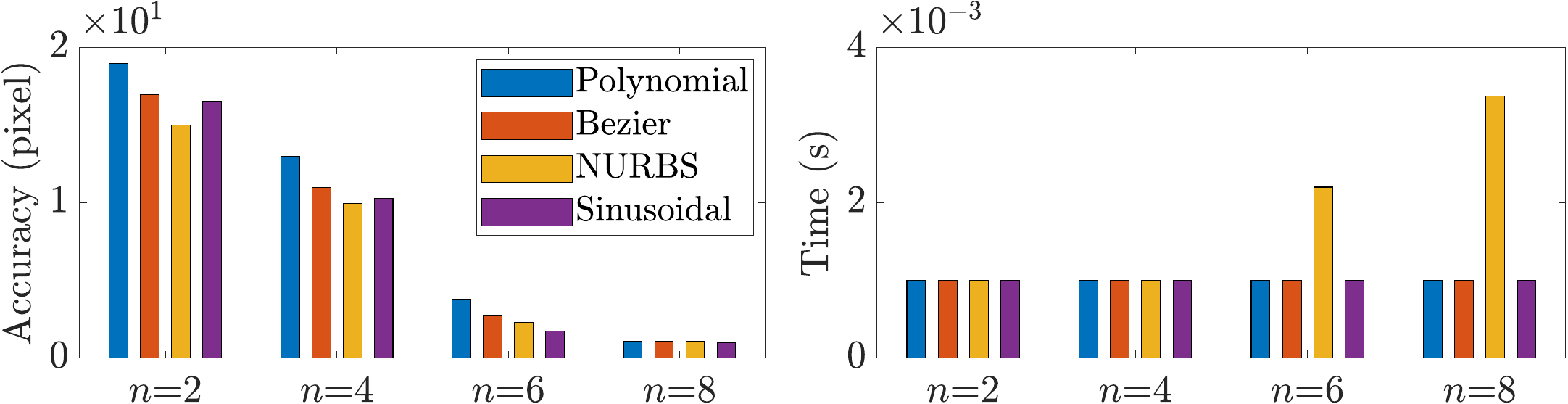}
	\vspace{-0.7cm}
	\caption{
	Comparison results among four regression methods among 10000 shape sets.
	The abscissa is the fitting order $n$.}
	\label{fig30}
\end{figure}

\vspace{-0.3cm}
\subsection{Comparison of the Feature Extraction Methods}\label{section4b}
In this section, 10000 sets of centerline observations are collected by commanding the robot to continuously deform the elastic rod, which are then used to evaluate the proposed parameterized regression features.
For that, we calculate the average error between the feedback shape $\bar{\mathbf{c}}$ and the reconstructed shape $\mathbf{B}\mathbf{s}$ as follows mean$(\|  \bar{\mathbf{c}} - \mathbf{Bs}\|)$.
The results also compare the computation time required for each method.

Fig. \ref{fig30} shows that as $n$ increases, the extraction works better, further helping represent the object's shape.
As for the same $n$, NURBS performs best, Sinusoidal and Bezier are similar, and Polynomial is prone to overfitting with higher $n$.
For NURBS, we set $\omega_j=0.327 ,j\in[0,n]$ as the optimal value.
Theoretically, the performance of NURBS can be improved by adjusting $\omega_j$.
In the level of time, NURBS is the slowest with the most iterative calculations, yet, it still enables to implement servo-loops of well above 30Hz.
Hence, throughout the rest of the experiments, we use NURBS (with $n=8$) to characterize and control the feedback shapes of the manipulated elastic object.

\begin{figure}[t]
	\vspace{-0.5cm}
	\centering
	\subfloat[RLS]{\includegraphics[scale=0.173]{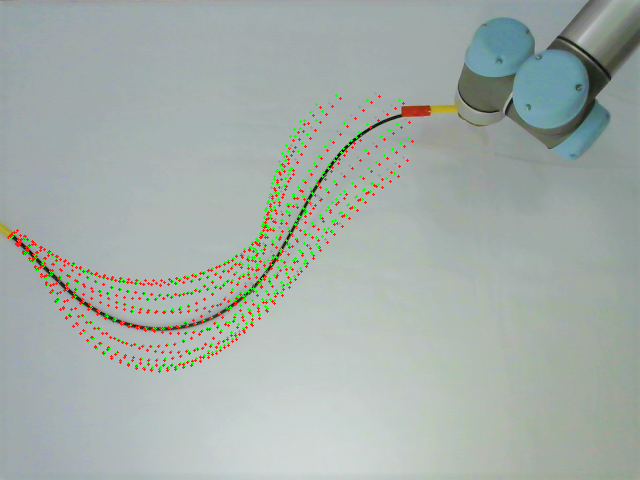}}
	\subfloat[LKF]{\includegraphics[scale=0.173]{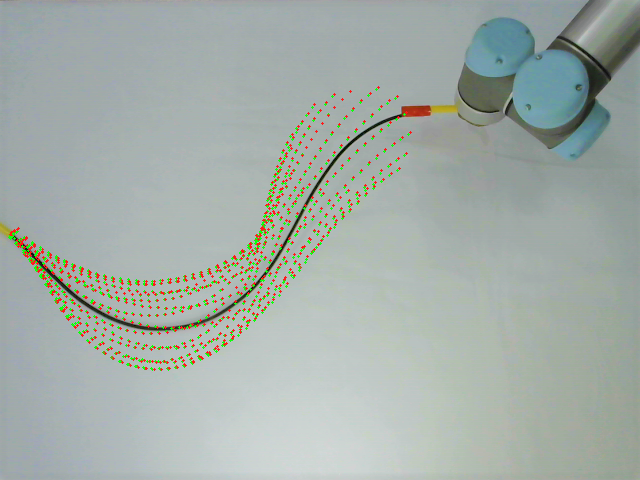}}
	\subfloat[UKF]{\includegraphics[scale=0.173]{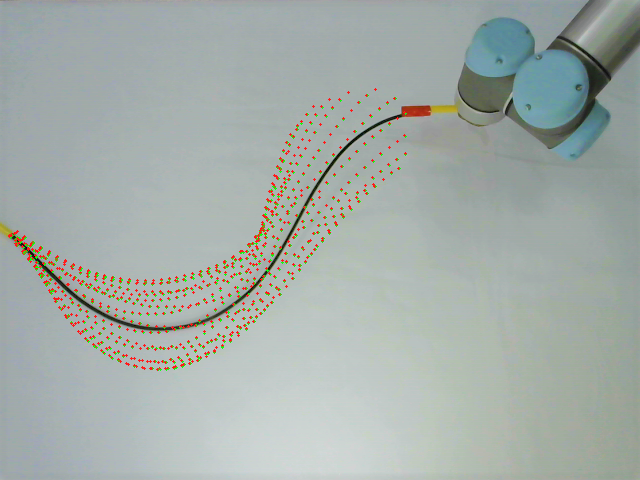}}
	\vspace{-0.16cm}
	\caption{
	Comparison between the visually measured shape (dashed green line) and its approximation shape with NURBS series (dashed red line).}
	\label{fig14}
\end{figure}

\begin{figure}[t]
	\centering
	\includegraphics[scale=0.25]{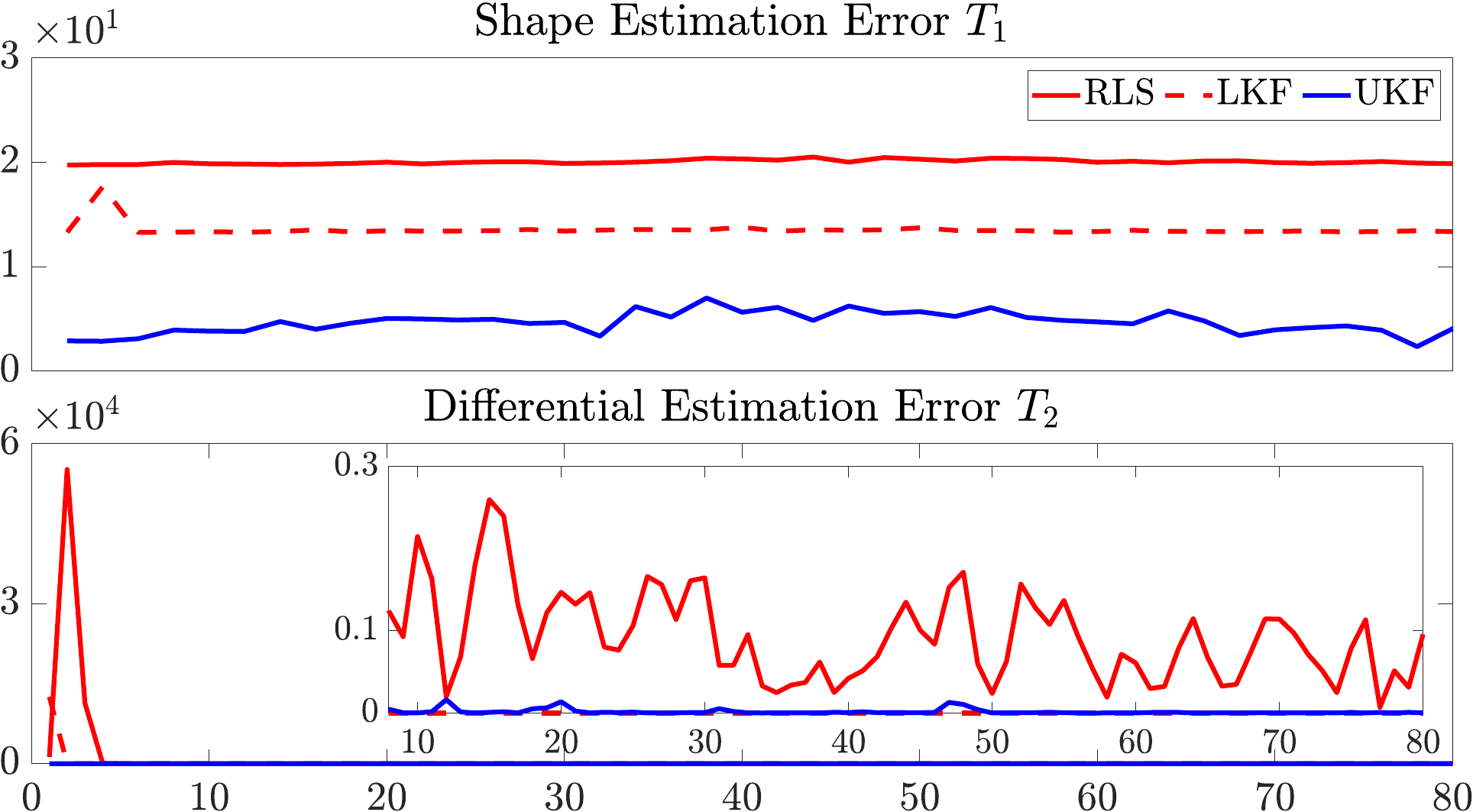}
	\vspace{-0.2cm}
	\caption{
	Profiles of $T_1$ and $T_2$ that are computed along the circular trajectory.
	The abscissa is the step size.}
	\label{fig15}
	\vspace{-0.47cm}
\end{figure}

\begin{figure*}
	\vspace{-0.8cm}
	\centering
	\subfloat[Experiment 1-RLS]
	{\includegraphics[scale=0.17]{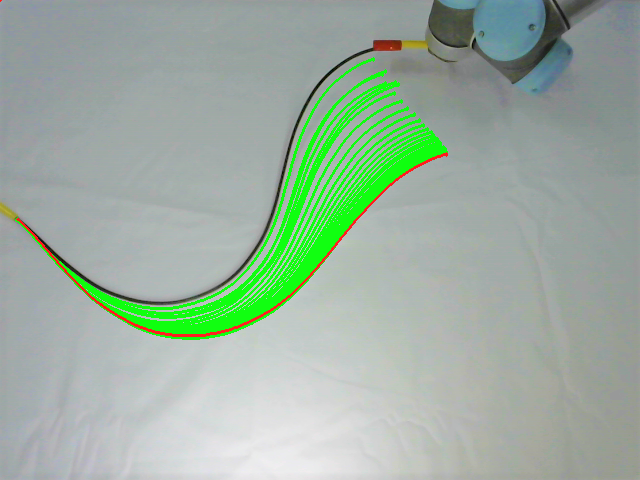}}
	\subfloat[Experiment 2-RLS]
	{\includegraphics[scale=0.17]{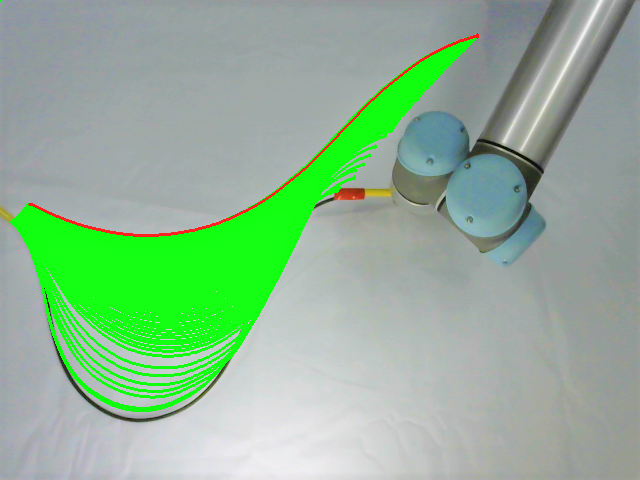}}
	\subfloat[Experiment 3-RLS]
	{\includegraphics[scale=0.17]{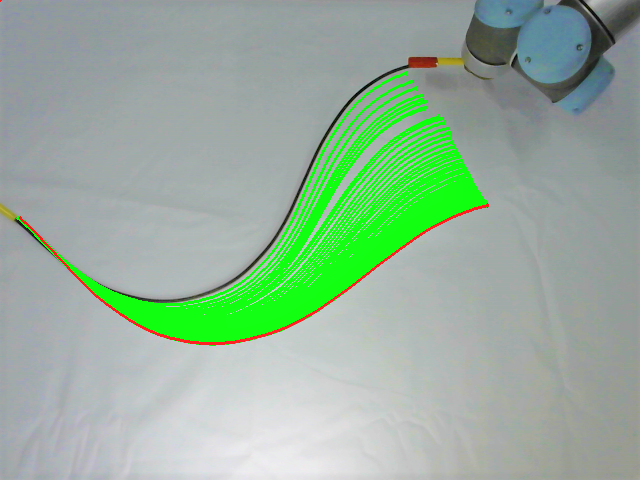}}
	\subfloat[Experiment 4-RLS]
	{\includegraphics[scale=0.17]{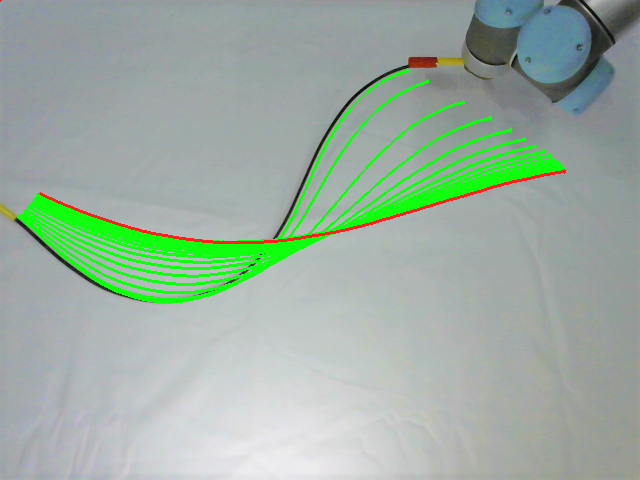}}
	\subfloat[Experiment 5-RLS]
	{\includegraphics[scale=0.17]{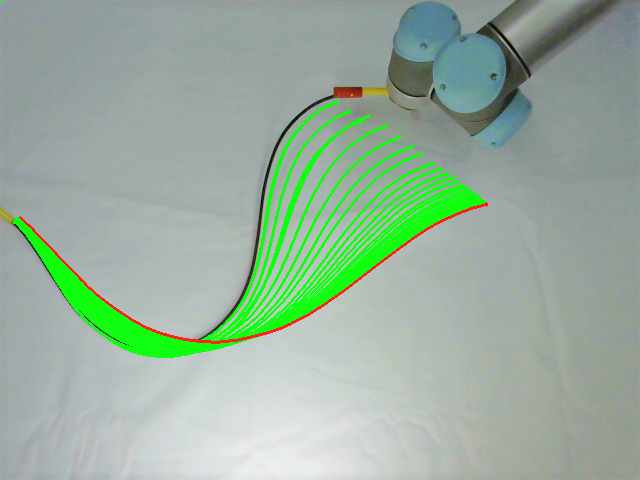}}
	\subfloat[Experiment 6-RLS]
	{\includegraphics[scale=0.17]{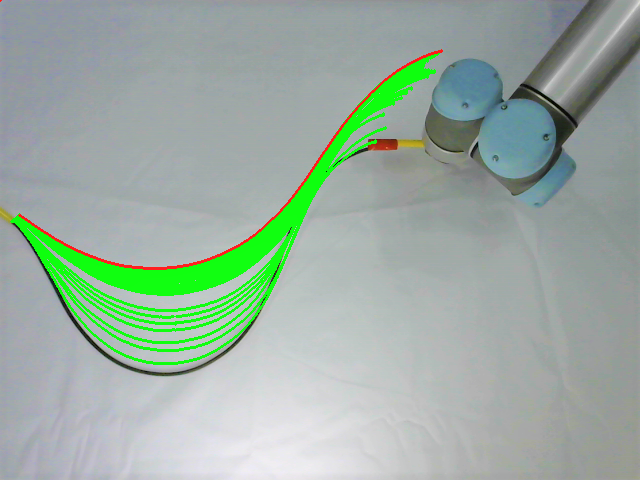}}

	\vspace{-0.35cm}
	\subfloat[Experiment 1-LKF]
	{\includegraphics[scale=0.17]{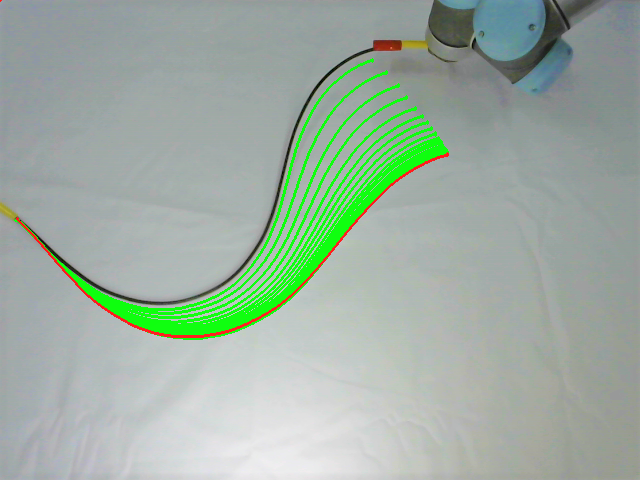}}
	\subfloat[Experiment 2-LKF]
	{\includegraphics[scale=0.17]{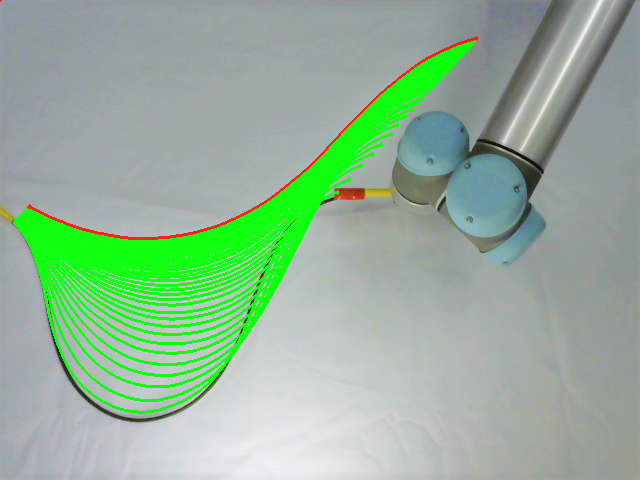}}
	\subfloat[Experiment 3-LKF]
	{\includegraphics[scale=0.17]{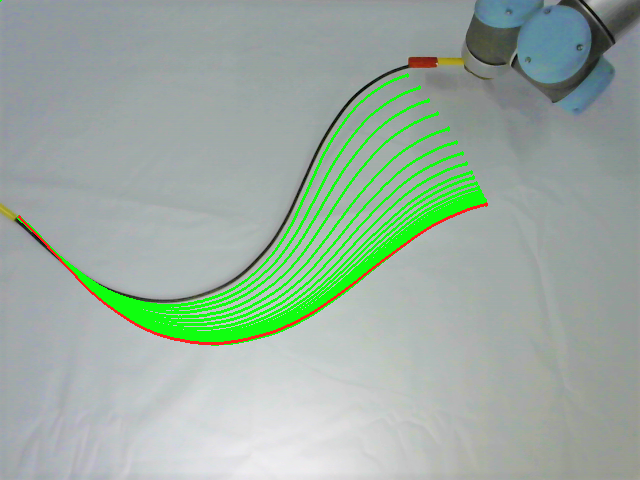}}
	\subfloat[Experiment 4-LKF]
	{\includegraphics[scale=0.17]{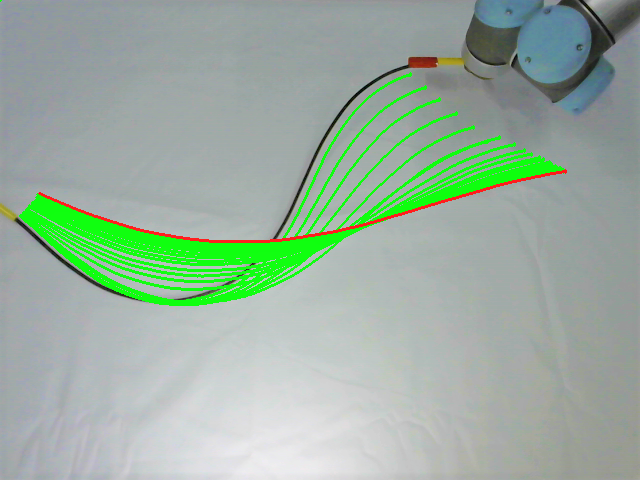}}
	\subfloat[Experiment 5-LKF]
	{\includegraphics[scale=0.17]{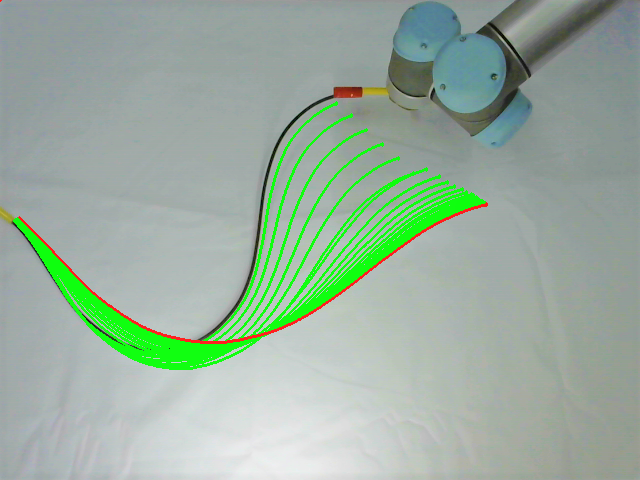}}
	\subfloat[Experiment 6-LKF]
	{\includegraphics[scale=0.17]{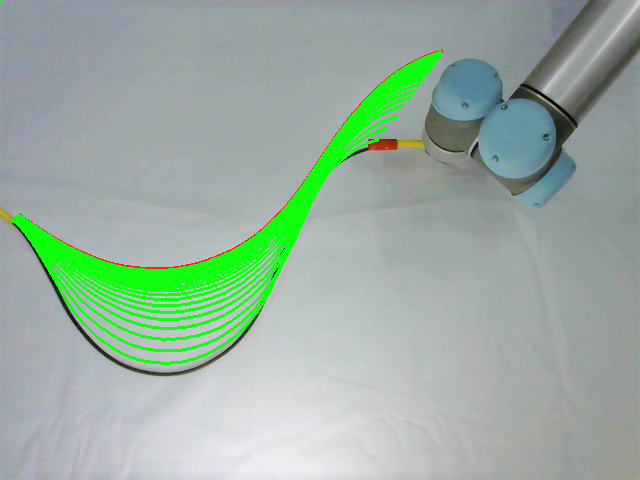}}

	\vspace{-0.35cm}
	\subfloat[Experiment 1-UKF]
	{\includegraphics[scale=0.17]{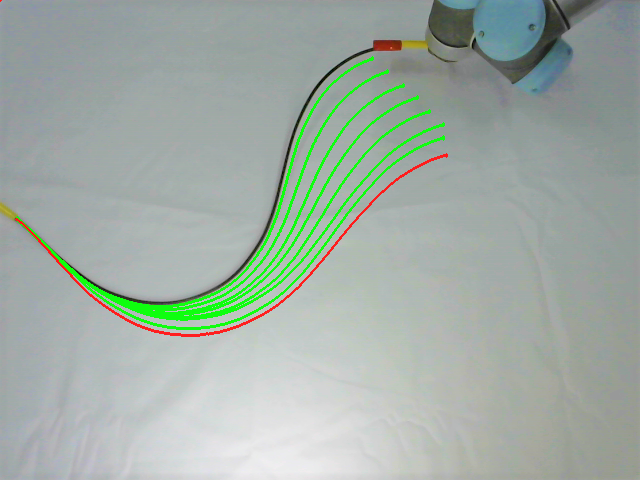}}
	\subfloat[Experiment 2-UKF]
	{\includegraphics[scale=0.17]{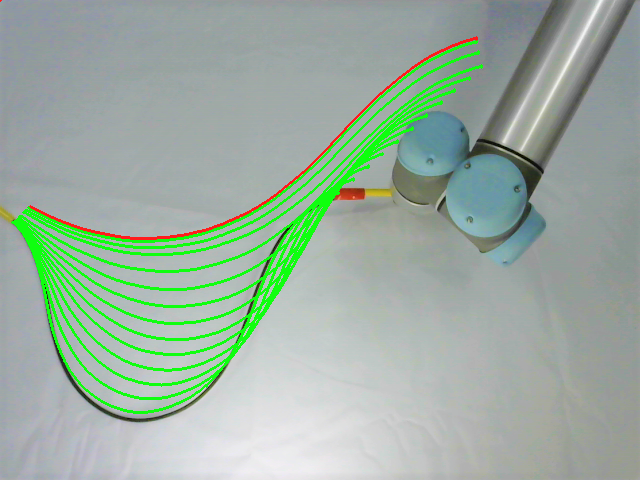}}
	\subfloat[Experiment 3-UKF]
	{\includegraphics[scale=0.17]{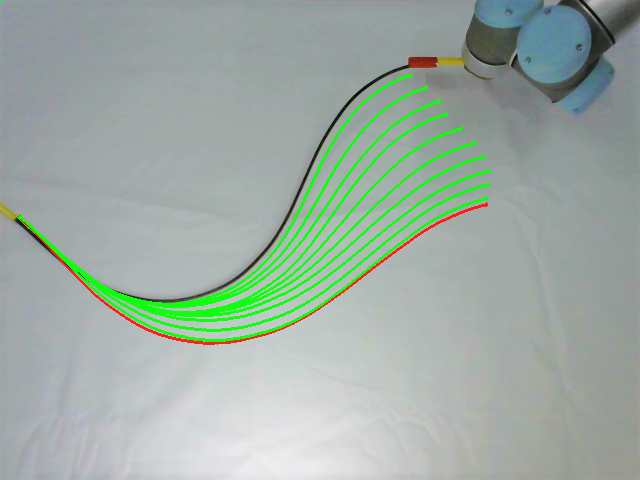}}
	\subfloat[Experiment 4-UKF]
	{\includegraphics[scale=0.17]{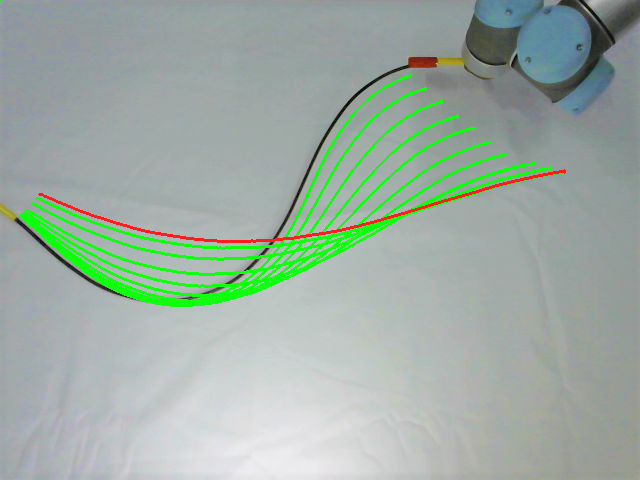}}
	\subfloat[Experiment 5-UKF]
	{\includegraphics[scale=0.17]{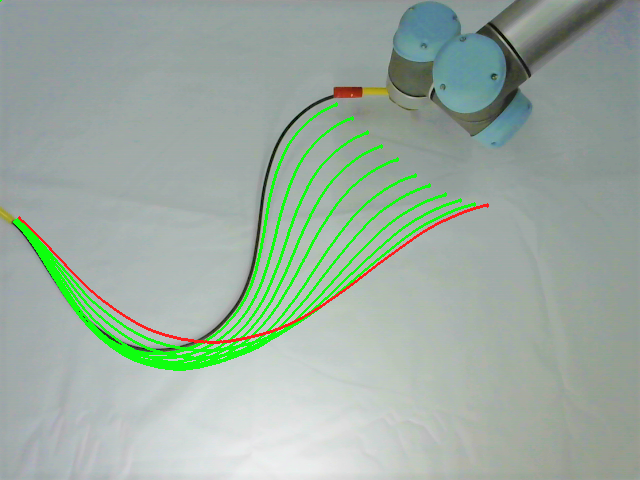}}
	\subfloat[Experiment 6-UKF]
	{\includegraphics[scale=0.17]{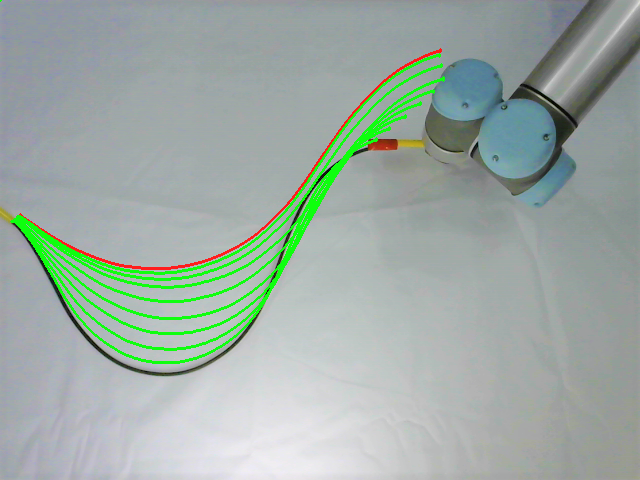}}

	\caption{
	Initial (solid black line), transition (solid green line) and target (solid red line) configurations in the six shape deformation experiments which have a variety of initial and target shape.
	These experiments are conducted within the motion controller \eqref{eq30} among RLS \cite{hosoda1994versatile}, LKF \cite{Qian2002Online} and UKF.}
	\label{fig16}
\end{figure*}

\begin{figure*}[htbp]
    \vspace{-0.6cm}
	\centering
	\subfloat[Experiment 1]{\includegraphics[scale=0.168]{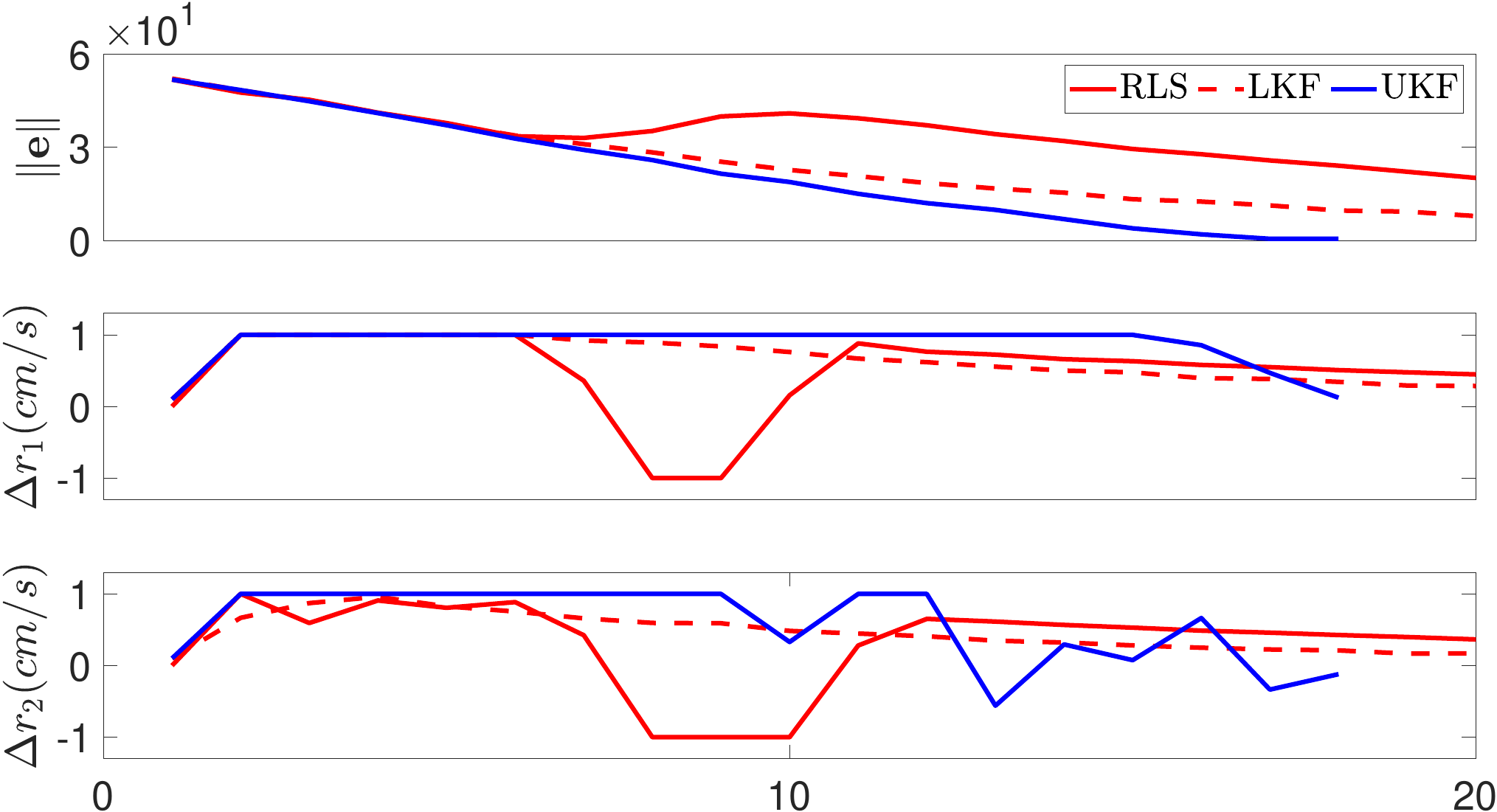}}
	\subfloat[Experiment 2]{\includegraphics[scale=0.168]{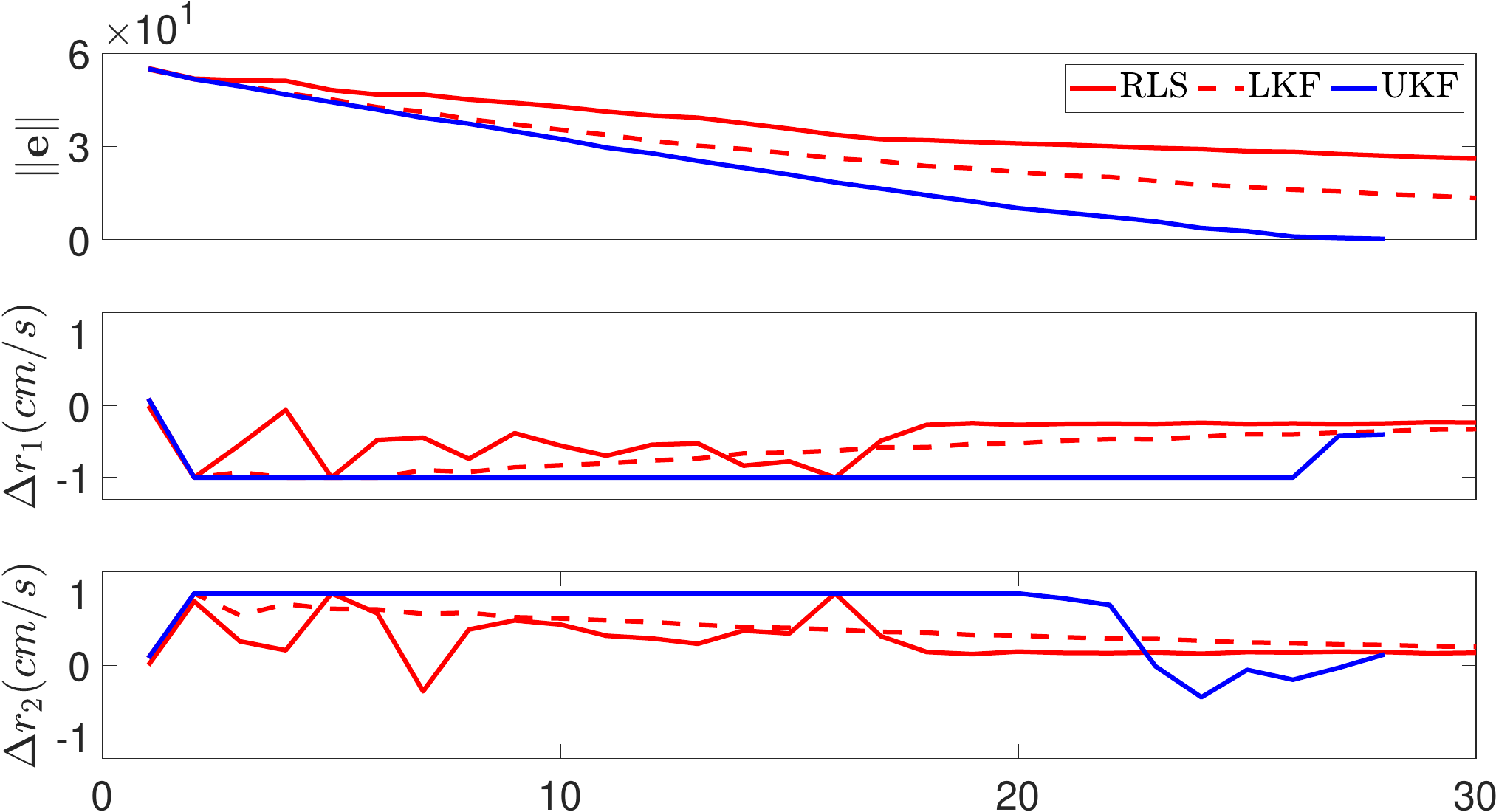}}
	\subfloat[Experiment 3]{\includegraphics[scale=0.168]{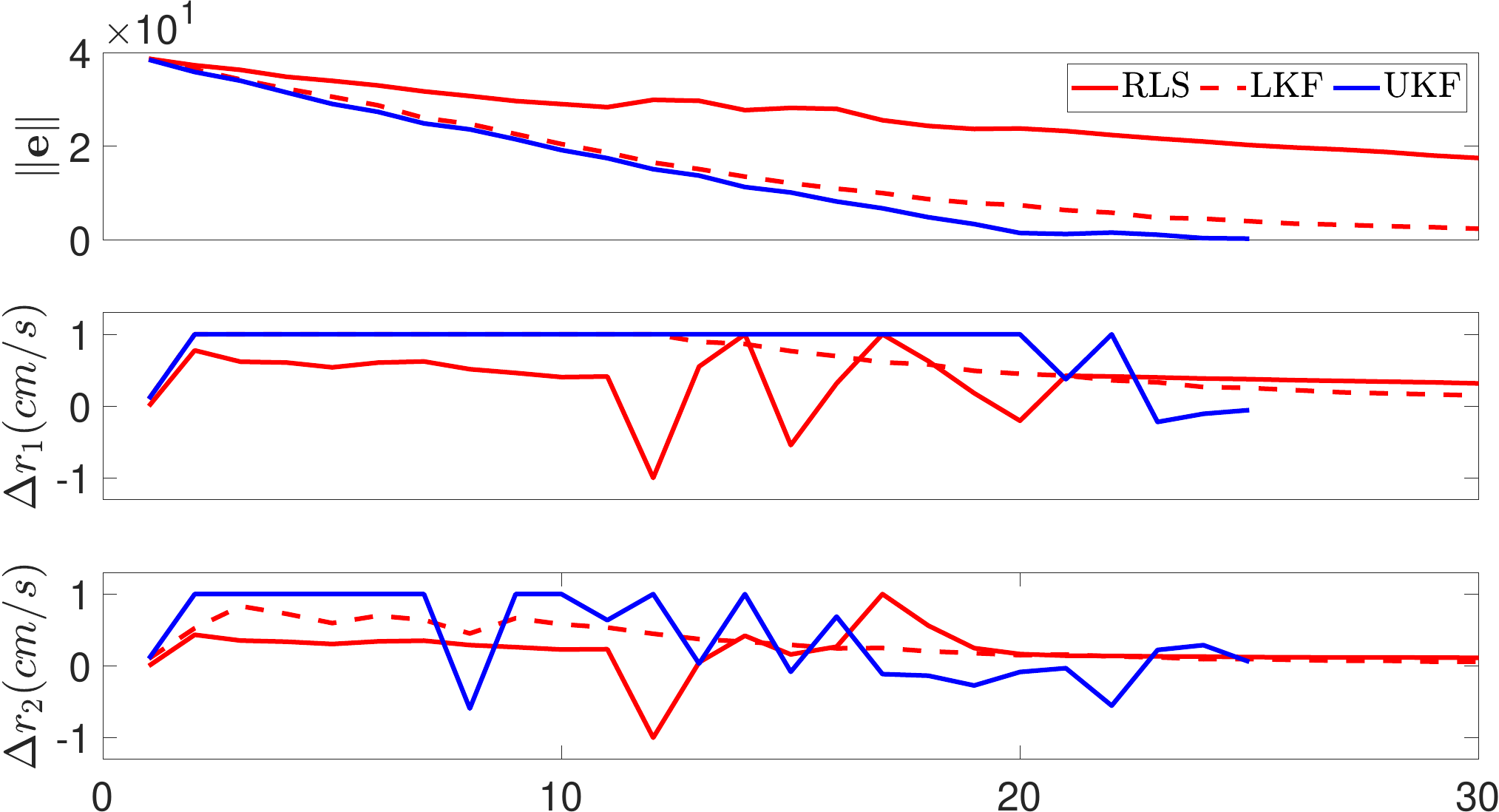}}
	
	\vspace{-0.3cm}
	\subfloat[Experiment 4]{\includegraphics[scale=0.168]{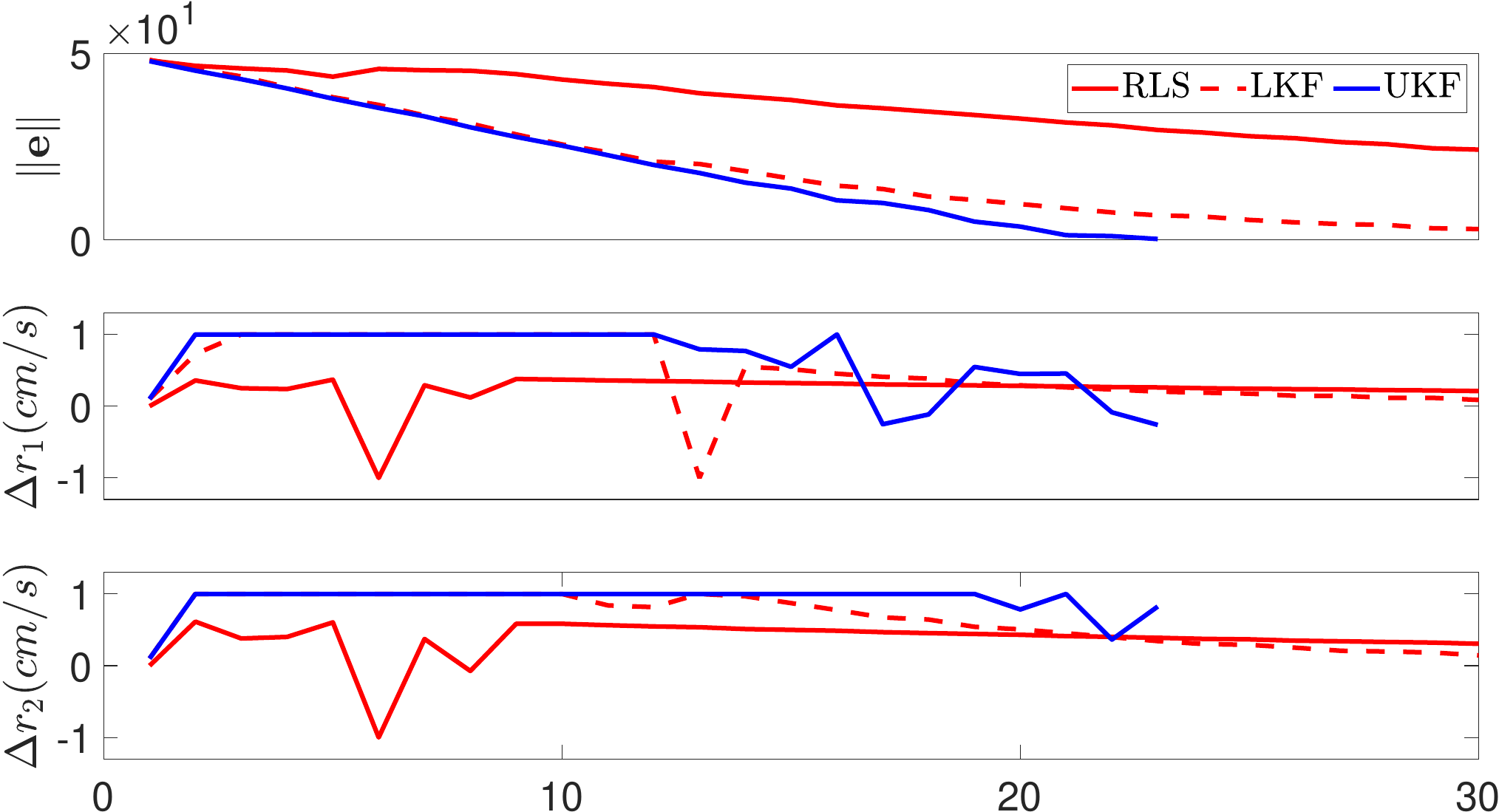}}
	\subfloat[Experiment 5]{\includegraphics[scale=0.168]{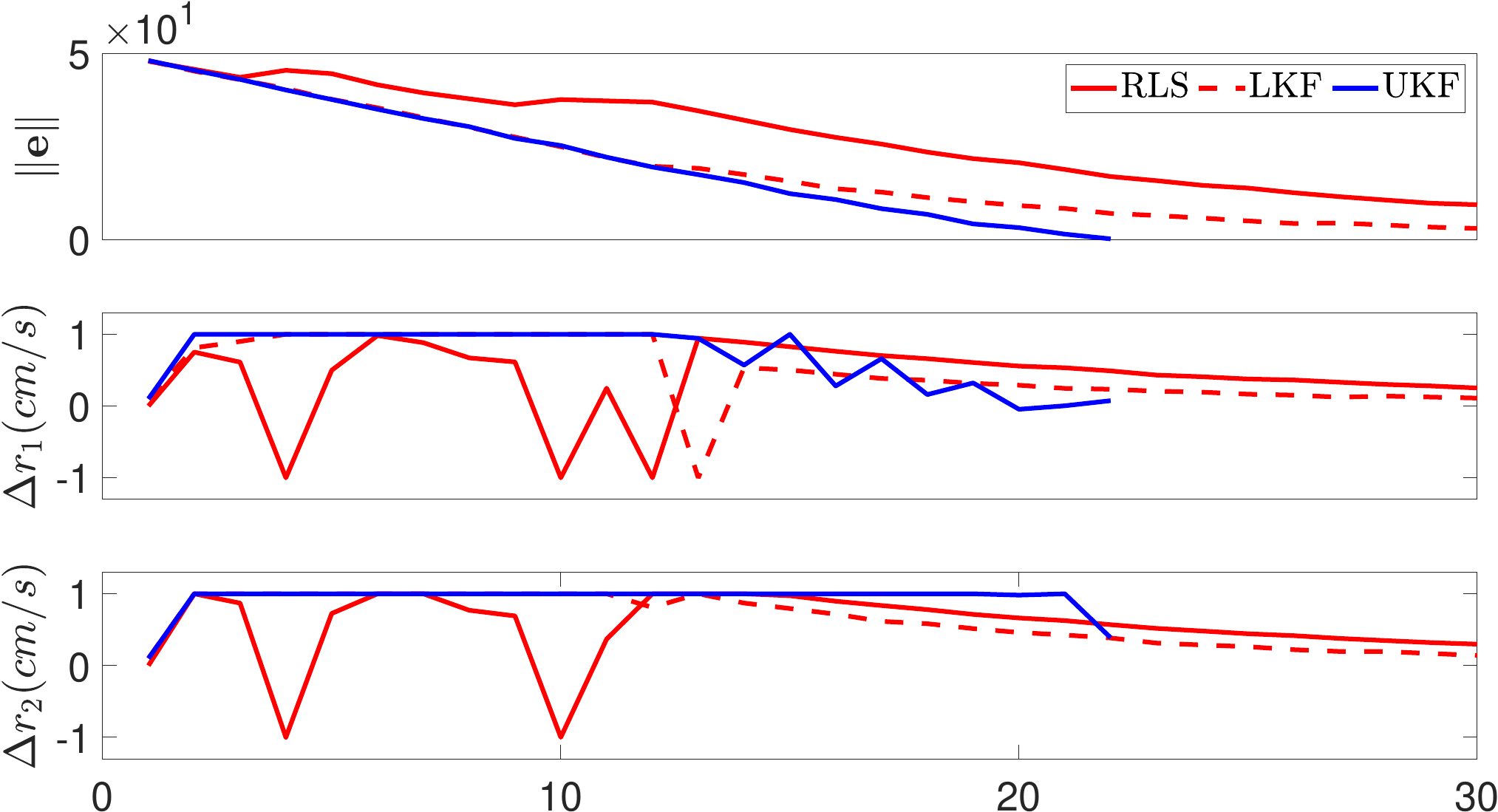}}
	\subfloat[Experiment 6]{\includegraphics[scale=0.168]{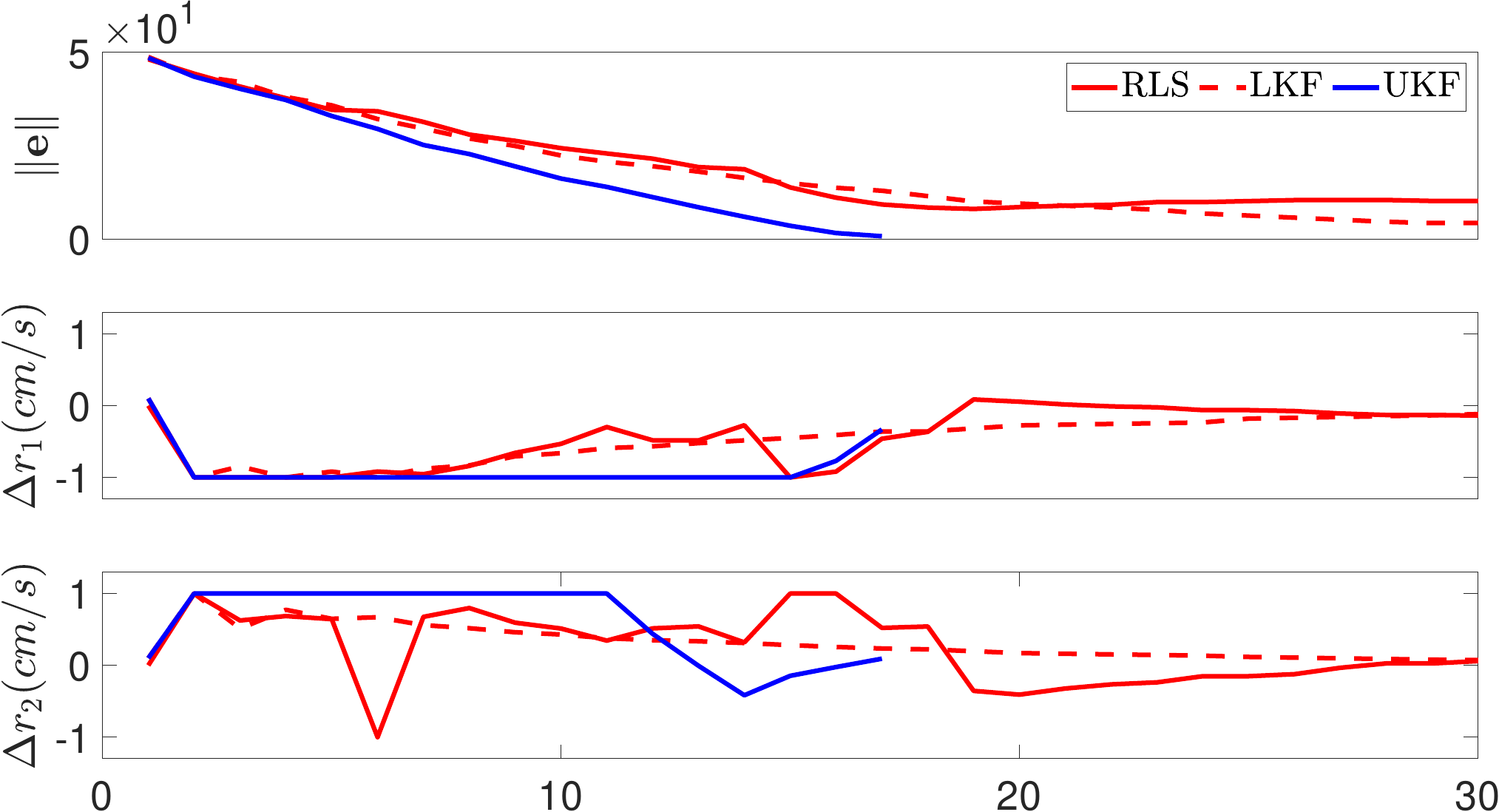}}
 
	\vspace{-0.16cm}
	\caption{
	Profiles of $\| \mathbf{e}_k \|$ and $\Delta \mathbf{r}_k$ among RLS, LKF and UKF within six shape deformation experiments.
	The abscissa is the step size.}
	\label{fig17}
\end{figure*}

\begin{figure}[htbp]
	\vspace{-0.6cm}
	\centering
	\subfloat[Fault case 1] {\includegraphics[scale=0.18]{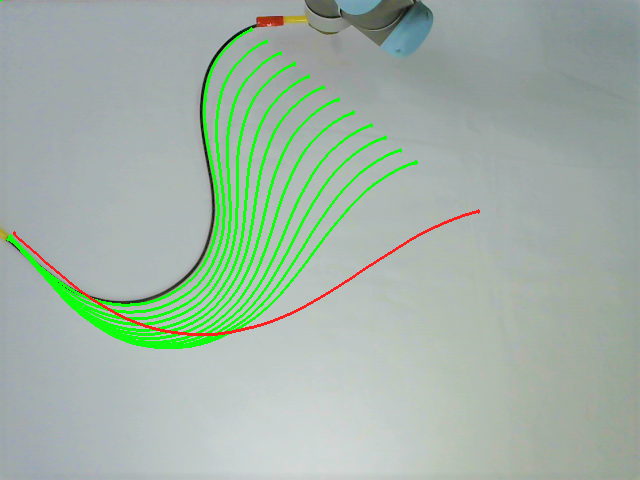}\label{fig20a}}
	\subfloat[Fault case 2]
	{\includegraphics[scale=0.18]{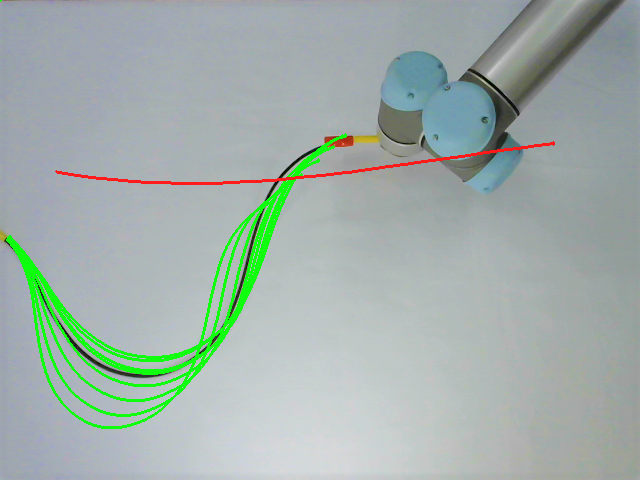}\label{fig20b}}
	\vspace{-0.21cm}
	\caption{Fault shape deformation experiments display.}
	\label{fig22}
	\vspace{-0.4cm}
\end{figure}

\begin{figure}[htbp]
	\vspace{-0.6cm}
	\centering
	\subfloat[FIX]  {\includegraphics[scale=0.125]{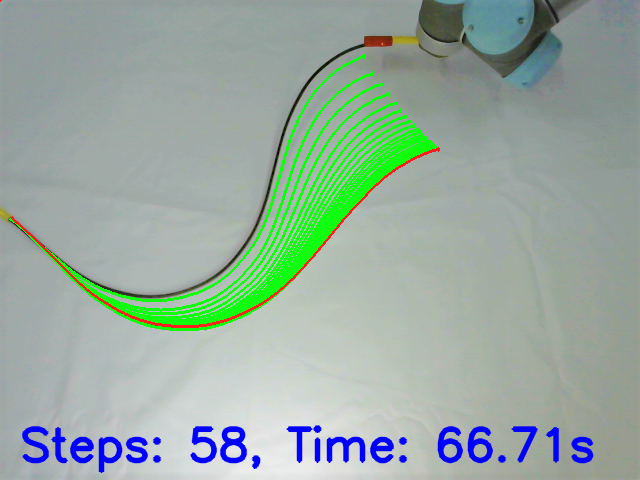}}
	\subfloat[ISE]  {\includegraphics[scale=0.125]{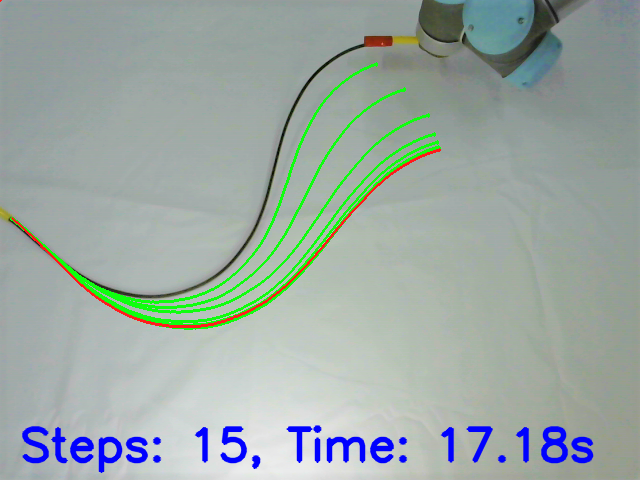}}
	\subfloat[IAE]  {\includegraphics[scale=0.125]{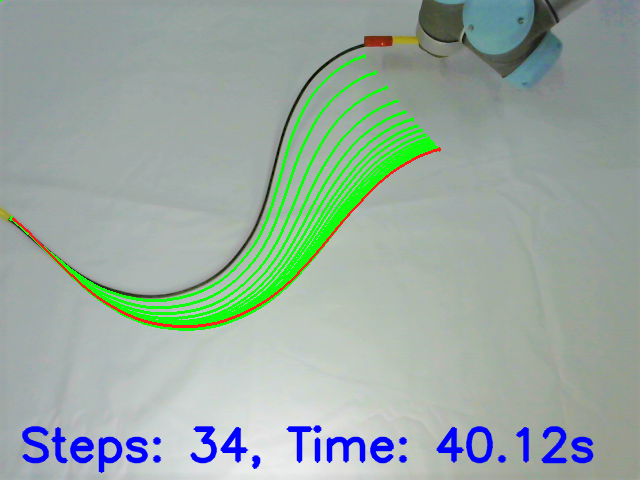}}
	
	\vspace{-0.3cm}
	\subfloat[JEU-1]{\includegraphics[scale=0.125]{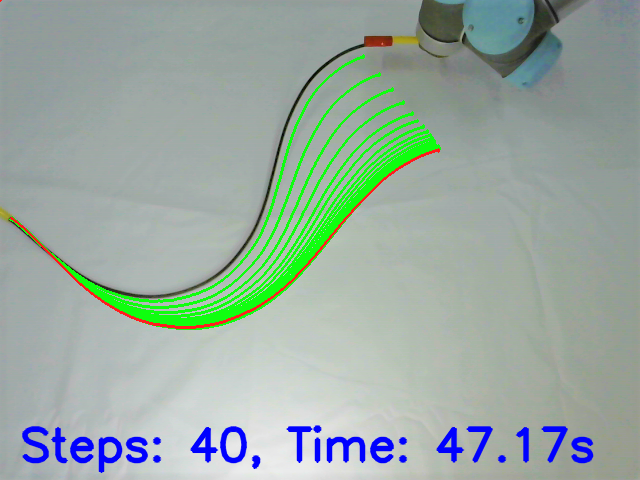}}
	\subfloat[JEU-2]{\includegraphics[scale=0.125]{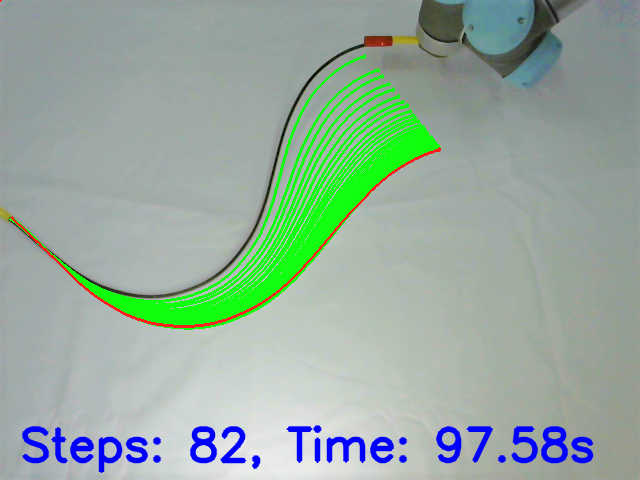}}
	\subfloat[JEU-3]{\includegraphics[scale=0.125]{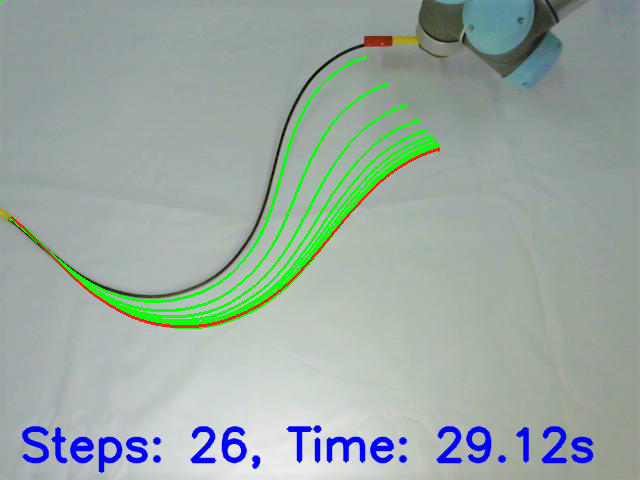}}
    \vspace{-0.16cm}
	\caption{
	Initial (solid black line), transition (solid green line) and target (solid red line) configurations among FIX, ISE, IAE,
	JEU-1 (${\omega}_1=0.5, {\omega}_2=0.5$), 
	JEU-2 (${\omega}_1=0.1, {\omega}_2=0.9$), and 
	JEU-3 (${\omega}_1=0.9, {\omega}_2=0.1$).
	The comparisons are conducted within LKF \cite{Qian2002Online}.}
	\label{fig18}
	\vspace{-0.6cm}
\end{figure}

\begin{figure}[htbp]
	\vspace{-0.2cm}
	\centering
	\includegraphics[scale=0.24]{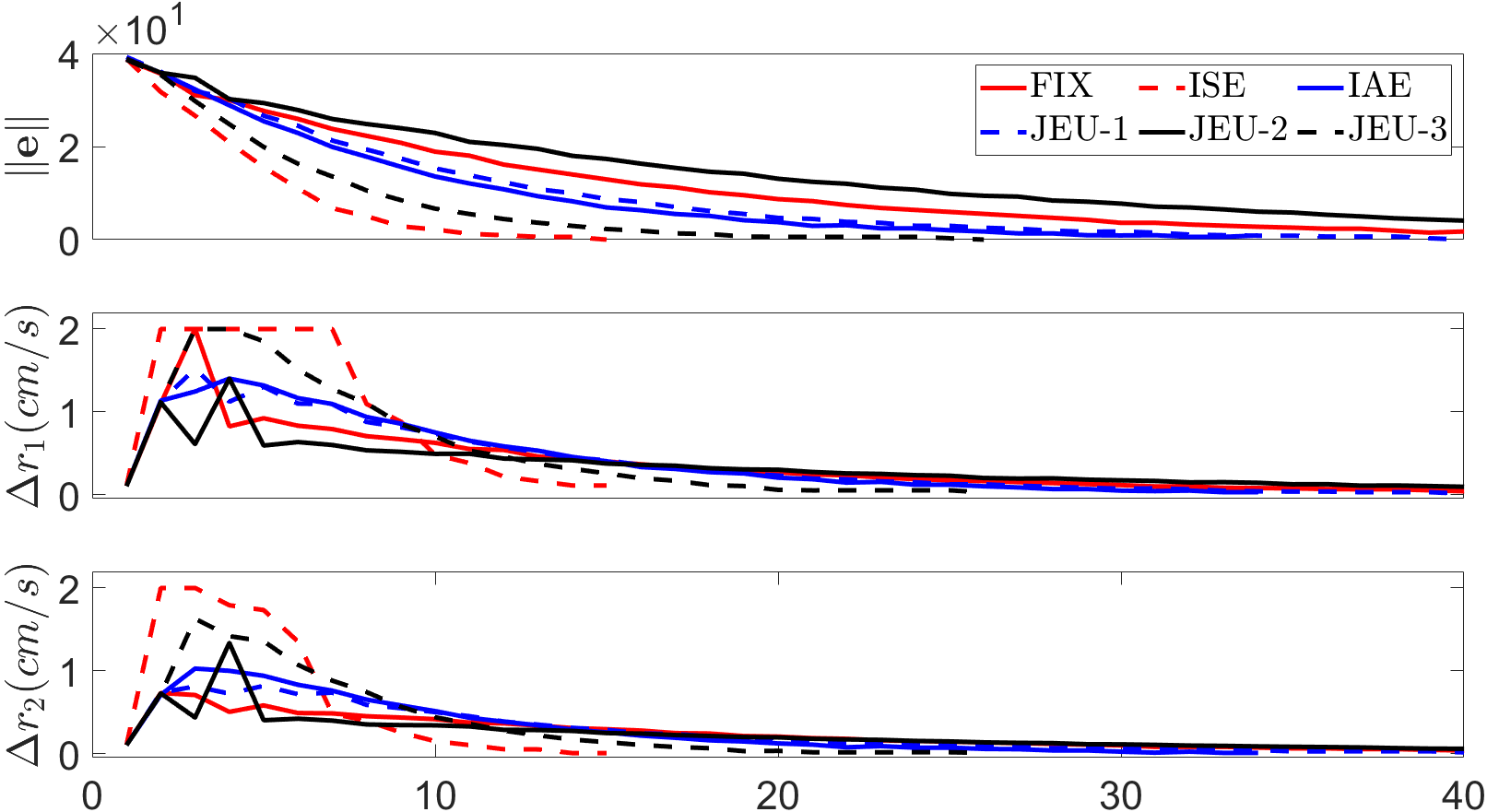}
	\vspace{-0.3cm}
	\caption{
	Profiles of $\|\mathbf{e}_k\|$ and $\Delta \mathbf{r}_k$ among FIX, ISE, IAE, JEU-1, JEU-2, and JEU-3.
	The abscissa is the step size.}
	\label{fig19}
\end{figure}


\vspace{-0.3cm}
\subsection{Estimation of the Deformation Jacobian Matrix}
In this section, we evaluate the performance (and thus, the suitability) of the UKF algorithm to estimate the system's differential model.
For that, the robot is commanded to move along a circular trajectory while rigidly grasping the elastic rod; 
These actions produce feedback shapes that progressively deform into a wide variety of configurations.
During these motions, we simultaneously approximate the model \eqref{eq-15} with 3 different methods: UKF, recursive least squares (RLS) \cite{hosoda1994versatile}, and linear Kalman filter (LKF) \cite{Qian2002Online}.
Fig. \ref{fig14} depicts three measured shapes (dashed green line) and their corresponding approximated shapes (dashed red line), all computed with NURBS.
The following two error metrics are used to evaluate the estimation algorithms:
\begin{align}
\label{eq24}
\begin{array}{*{20}{c}}
{{T_1} = \| {\widehat{\bar{\mathbf{c}}}_k - \bar{\mathbf{c}}_k} \|},&{{T_2} = \| {{\Delta \mathbf{s}_k} - {{\widehat{\mathbf{J}}}_k}\Delta {\mathbf{r}_k}} \|}
\end{array}
\end{align}
where $\widehat{\bar{\mathbf{c}}}_k = \mathbf{B}\widehat{\mathbf{s}}_k$ is the approximated shape, with
$\widehat{\mathbf{s}}_k$ calculated as $\widehat{\mathbf{s}}_k = \widehat{\mathbf{s}}_{k-1} + \widehat{\mathbf{J}}_k \Delta \mathbf{r}_k$, with $\widehat{\mathbf{s}}_0 = \mathbf{s}_0$ for $\widehat{\mathbf{s}}_0$ and $\mathbf{s}_0$ as the initial values of $\widehat{\mathbf{s}}_k$ and $\mathbf{s}_k$, respectively.
Fig. \ref{fig15} shows the evolution of $T_1$ and $T_2$ during the manipulation motions.
Compared with RLS and LKF, we can see that UKF has a higher accuracy in the estimation of the Jacobian matrix, with both errors rapidly converging to a small steady error.
The lack of noticeable fluctuations in the UKF-computed plots reflects the adaptability of the algorithm to different local regions.


\vspace{-0.3cm}
\subsection{Shape Servoing of Elastic Rods}
In this section, six shape control experiments with a variety of initial and desired configurations are conducted with the proposed controller \eqref{eq30} with the fixed gain ($\lambda=2.8$), combined with RLS \cite{hosoda1994versatile}, LKF \cite{Qian2002Online}, and UKF.
The target shapes are obtained by driving the robot to an arbitrary position and recording the corresponding feature vector $\mathbf s^*$ (this ensures the reachability of the task).
No other information (e.g., the corresponding robot pose $\mathbf r$) is used in these control experiments where the robot automatically drives the cable to the desired configuration, by using visual feedback only.
In these tests, the following safety velocity limits are used for each motion coordinate of the robot: $\|\Delta {r}_k\|\le0.01$ m/s.

Fig. \ref{fig16} depicts the active shaping motions (solid green lines) of the elastic rod towards the desired shape (solid red line) of each control experiment with the three estimation algorithms.
These visually-guided motions demonstrate that the proposed framework (representation, approximation, and control) can automatically drive the object to different shape configurations.
Fig. \ref{fig17} shows the profiles of $\mathbf{e}_k$ and $\Delta \mathbf{r}_k$ that were computed during the experiments. 
The figures show that UKF presents the best performance and RLS the worst.
UKF can accurately estimate the Jacobian matrix, which helps to drive the robot $\Delta \mathbf r_k$ in the correct directions that minimize the vision error $\|\mathbf e_k\|$.
Fig. \ref{fig16} also shows that NURBS can adequately approximate the object's centerline.
These results demonstrate the controller can guide the robot to deform the objects into multiples shapes without damaging it (i.e., over-stretching, or over-compressing).


To thoroughly test the performance of the proposed method, we also tested the following two fault conditions were the robot cannot perform the shape servoing task (shown in Fig. \ref{fig22}):
1) If estimated Jacobian matrix has not be properly and sufficiently initialized at the time instant zero $\widehat{\mathbf J}_0$ (see Remark 2), the motion controller may drive the robot in the wrong directions.
2) If the target feature vector $\mathbf s^*$ represents an unfeasible shape (i.e., a configuration that cannot be achieved with the current object-manipulation setup), the controller cannot drive the object towards it, and it will typically reach a local minimum.


\vspace{-0.3cm}
\subsection{Parameter Optimization Comparison}\label{section4e}
In this section, the performance of the parameter optimization criteria (ISE \eqref{eq-39}, IAE \eqref{eq-42} and JEU \eqref{eq-44}) is compared with the traditional fixed gain (FIX) control method (i.e., for the case where $\lambda=7.2$ is constant).
To evaluate the impact of the weights in JEU (\ref{eq-44}), the following three sets of parameters are used:
1) JEU-1: ${\omega}_1=0.5,{\omega}_2=0.5$, in this case, the effect is similar to IAE;
2) JEU-2: ${\omega}_1=0.1, {\omega}_2=0.9$, which pays more attention to the smoothness of $\Delta \mathbf{r}_k$;
3) JEU-3: ${\omega}_1=0.9, {\omega}_2=0.1$, which improves the convergence of $\mathbf{e}_k$;
The adaptive gain is initialized as follows $\lambda_0 = 7.2$ and is updated with a learning rate gain $d=0.001$.
In these experiments, the upper limit of the velocity command for each motion coordinate of the robot is set to $\|\Delta {r}_k\| \le0.02$ m/s.


Fig. \ref{fig18} presents performance metrics obtained with the different parameter optimization methods.
The table shows that ISE has the shortest convergence time, followed by JEU-3, while JEU-2 is the slowest.
As ISE focuses on compensating errors, the resulting motion command $\Delta \mathbf{r}_k$ presents larger  fluctuations. 
Fig. \ref{fig19} shows that JEU-2 provides slower convergence with smoother profiles $\Delta \mathbf{r}_k$ than JEU-3; The controls $\Delta \mathbf{r}_k$ computed with JEU-3 are relatively larger and close to saturation.
This shows that by varying the weights of JEU, we can tune the system to match various target performances.
The above analyses demonstrate that the proposed parameter optimization criteria can effectively adjust ${\lambda}$ according to different task requirements.


\vspace{-0.3cm}
\section{Conclusion}\label{sect6}
This paper presents a shape servoing framework for automatically manipulating an elastic rod to desired configurations.
A general method based on parametric features (e.g. sinusoisal, polynomial, B$\acute{\text{e}}$zier, and NURBS) is presented to characterize the shape of the object with a compact feedback-like vector (on which an explicit servo-loop is established upon).
A UKF-based online estimator is proposed to coordinate the driving motions of the robot with the produced visual measurements.
An adaptive velocity controller is derived considering various performance criteria. 
The stability of the system is rigorously analyzed using Lyapunov theory.
A detailed experimental study is presented to validate the effectiveness of the new control method.

The proposed method has some limitations.
First, the proposed framework is only suitable to manipulate mostly elastic materials (i.e., those whose shape is mainly determined by its potential energy); The method does not consider inelastic or non-homogeneous materials.
Second, the method cannot judge if the desired configuration is reachable, which can lead to failure during the task execution.
Third, the method requires a full (i.e., un-occluded) visual observation of the object at all times.
Future work includes the design of similar control schemes which can handle occlusions of the object; Our team is currently developing a neural network which can predict the shape of the object based on partial observations.

\appendices
\ifCLASSOPTIONcaptionsoff
\newpage
\fi

\bibliography{biblio}

\begin{thebibliography}{10}
\providecommand{\url}[1]{#1}
\csname url@samestyle\endcsname
\providecommand{\newblock}{\relax}
\providecommand{\bibinfo}[2]{#2}
\providecommand{\BIBentrySTDinterwordspacing}{\spaceskip=0pt\relax}
\providecommand{\BIBentryALTinterwordstretchfactor}{4}
\providecommand{\BIBentryALTinterwordspacing}{\spaceskip=\fontdimen2\font plus
\BIBentryALTinterwordstretchfactor\fontdimen3\font minus
  \fontdimen4\font\relax}
\providecommand{\BIBforeignlanguage}[2]{{%
\expandafter\ifx\csname l@#1\endcsname\relax
\typeout{** WARNING: IEEEtran.bst: No hyphenation pattern has been}%
\typeout{** loaded for the language `#1'. Using the pattern for}%
\typeout{** the default language instead.}%
\else
\language=\csname l@#1\endcsname
\fi
#2}}
\providecommand{\BIBdecl}{\relax}
\BIBdecl

\bibitem{yu2022global}
M.~Yu, K.~Lv, H.~Zhong, S.~Song, and X.~Li, ``Global model learning for large
  deformation control of elastic deformable linear objects: An efficient and
  adaptive approach,'' \emph{IEEE Transactions on Robotics}, vol.~39, no.~1,
  pp. 417--436, 2022.

\bibitem{huang2022task}
J.~Huang and K.~S. Au, ``Task-oriented grasping position selection in
  deformable object manipulation,'' \emph{IEEE Robotics and Automation
  Letters}, vol.~8, no.~2, pp. 776--783, 2022.

\bibitem{cherubini2020model}
A.~Cherubini, V.~Ortenzi, A.~Cosgun, R.~Lee, and P.~Corke, ``Model-free
  vision-based shaping of deformable plastic materials,'' \emph{The
  International Journal of Robotics Research}, vol.~39, no.~14, pp. 1739--1759,
  2020.

\bibitem{zhang2023visual}
F.~Zhang and Y.~Demiris, ``Visual-tactile learning of garment unfolding for
  robot-assisted dressing,'' \emph{IEEE Robotics and Automation Letters}, 2023.

\bibitem{qin2023dual}
Y.~Qin, A.~Escande \emph{et~al.}, ``Dual-arm mobile manipulation planning of a
  long deformable object in industrial installation,'' \emph{IEEE Robotics and
  Automation Letters}, vol.~8, no.~5, pp. 3039--3046, 2023.

\bibitem{Journals:Sanchez2018}
J.~Sanchez, J.-A. Corrales, B.-C. Bouzgarrou, and Y.~Mezouar, ``Robotic
  manipulation and sensing of deformable objects in domestic and industrial
  applications: a survey,'' \emph{{Int. J. Robot. Res.}}, vol.~37, no.~7, pp.
  688--716, 2018.

\bibitem{navarro2017fourier}
D.~Navarro-Alarcon and Y.-H. Liu, ``Fourier-based shape servoing: a new
  feedback method to actively deform soft objects into desired 2-d image
  contours,'' \emph{IEEE Transactions on Robotics}, vol.~34, no.~1, pp.
  272--279, 2018.

\bibitem{navarro2014visual}
D.~Navarro-Alarcon, Y.-h. Liu, J.~G. Romero, and P.~Li, ``On the visual
  deformation servoing of compliant objects: Uncalibrated control methods and
  experiments,'' \emph{The International Journal of Robotics Research},
  vol.~33, no.~11, pp. 1462--1480, 2014.

\bibitem{1321161}
F.~{Chaumette}, ``Image moments: a general and useful set of features for
  visual servoing,'' \emph{IEEE Transactions on Robotics}, vol.~20, no.~4, pp.
  713--723, 2004.

\bibitem{Hu20193}
Z.~Hu, T.~Han, P.~Sun, J.~Pan, and D.~Manocha, ``3-d deformable object
  manipulation using deep neural networks,'' \emph{{IEEE Robot. Autom. Lett.}},
  vol.~4, no.~4, 2019.

\bibitem{laranjeira2017catenary}
M.~Laranjeira, C.~Dune, and V.~Hugel, ``Catenary-based visual servoing for
  tethered robots,'' in \emph{2017 IEEE International Conference on Robotics
  and Automation (ICRA)}.\hskip 1em plus 0.5em minus 0.4em\relax IEEE, 2017,
  pp. 732--738.

\bibitem{nair2017combining}
A.~Nair, D.~Chen, P.~Agrawal, P.~Isola, P.~Abbeel, J.~Malik, and S.~Levine,
  ``Combining self-supervised learning and imitation for vision-based rope
  manipulation,'' in \emph{2017 IEEE international conference on robotics and
  automation (ICRA)}.\hskip 1em plus 0.5em minus 0.4em\relax IEEE, 2017, pp.
  2146--2153.

\bibitem{9623343}
L.~Hu, D.~Navarro-Alarcon, A.~Cherubini, M.~Li, and L.~Li, ``On radiation-based
  thermal servoing: New models, controls, and experiments,'' \emph{IEEE
  Transactions on Robotics}, vol.~38, no.~3, pp. 1945--1958, 2022.

\bibitem{navarro2013model}
D.~Navarro-Alarcon, Y.-H. Liu \emph{et~al.}, ``Model-free visually servoed
  deformation control of elastic objects by robot manipulators,'' \emph{IEEE
  Transactions on Robotics}, vol.~29, no.~6, pp. 1457--1468, 2013.

\bibitem{alambeigi2018autonomous}
F.~Alambeigi, Z.~Wang, R.~Hegeman, Y.-H. Liu, and M.~Armand, ``Autonomous
  data-driven manipulation of unknown anisotropic deformable tissues using
  unmodelled continuum manipulators,'' \emph{IEEE Robotics and Automation
  Letters}, vol.~4, no.~2, pp. 254--261, 2018.

\bibitem{Qian2002Online}
J.~Qian and J.~Su, ``Online estimation of image jacobian matrix by kalman-bucy
  filter for uncalibrated stereo vision feedback,'' in \emph{IEEE International
  Conference on Robotics Automation}, 2002.

\bibitem{ibrahim2019optimal}
M.~A. Ibrahim, A.~K. Mahmood, and N.~S. Sultan, ``Optimal pid controller of a
  brushless dc motor using genetic algorithm,'' \emph{Int J Pow Elec \& Dri
  Syst ISSN}, vol. 2088, no. 8694, p. 8694, 2019.

\bibitem{kamal2014speed}
M.~M. Kamal, L.~Mathew, and S.~Chatterji, ``Speed control of brushless dc motor
  using intelligent controllers,'' in \emph{2014 Students Conference on
  Engineering and Systems}.\hskip 1em plus 0.5em minus 0.4em\relax IEEE, 2014,
  pp. 1--5.

\bibitem{huang2021non}
J.~Huang, Y.~Cai, X.~Chu, R.~H. Taylor, and K.~S. Au, ``Non-fixed contact
  manipulation control framework for deformable objects with active contact
  adjustment,'' \emph{IEEE Robotics and Automation Letters}, vol.~6, no.~2, pp.
  2878--2885, 2021.

\bibitem{Journals:Wang2018}
H.~{Wang}, B.~{Yang}, J.~{Wang}, X.~{Liang}, W.~{Chen}, and Y.~{Liu},
  ``Adaptive visual servoing of contour features,'' \emph{{IEEE/ASME} Trans. on
  Mechatronics}, vol.~23, no.~2, pp. 811--822, 2018.

\bibitem{piegl2012nurbs}
L.~Piegl and W.~Tiller, \emph{The NURBS book}.\hskip 1em plus 0.5em minus
  0.4em\relax Springer Science \& Business Media, 2012.

\bibitem{powell1981approximation}
M.~J.~D. Powell \emph{et~al.}, \emph{Approximation theory and methods}.\hskip
  1em plus 0.5em minus 0.4em\relax Cambridge university press, 1981.

\bibitem{xiong2006performance}
K.~Xiong, H.~Zhang, and C.~Chan, ``Performance evaluation of ukf-based
  nonlinear filtering,'' \emph{Automatica}, vol.~42, no.~2, pp. 261--270, 2006.

\bibitem{liu2023image}
J.~Liu, J.~Gao, W.~Yan, Y.~Chen, and B.~Yang, ``Image-based visual servoing of
  underwater vehicles for tracking a moving target using model predictive
  control with motion estimation,'' \emph{International Journal of Vehicle
  Design}, vol.~91, no. 1-3, pp. 46--66, 2023.

\bibitem{qi2021contour}
J.~Qi, G.~Ma, J.~Zhu, P.~Zhou, Y.~Lyu, H.~Zhang, and D.~Navarro-Alarcon,
  ``Contour moments based manipulation of composite rigid-deformable objects
  with finite time model estimation and shape/position control,''
  \emph{IEEE/ASME Transactions on Mechatronics}, vol.~27, no.~5, pp.
  2985--2996, 2021.

\bibitem{Zhuang1993Automatic}
M.~Zhuang and D.~P. Atherton, ``Automatic tuning of optimum pid controllers,''
  \emph{Control Theory Applications Iee Proceedings D}, vol. 140, no.~3, pp.
  216--224, 1993.

\bibitem{Xiong2005Study}
W.~Xiong, B.~Xu, and Q.~Zhou, ``Study on optimization of pid parameter based on
  improved pso,'' \emph{Computer Engineering}, vol.~31, no.~24, pp. 41--43,
  2005.

\bibitem{Sarpturk1987On}
S.~Sarpturk, Y.~Istefanopulos, and O.~Kaynak, ``On the stability of
  discrete-time sliding mode control systems,'' \emph{IEEE Transactions on
  Automatic Control}, vol.~32, no.~10, pp. 930--932, 1987.

\bibitem{ma2014adaptive}
J.~Ma, S.~S. Ge, Z.~Zheng, and D.~Hu, ``Adaptive nn control of a class of
  nonlinear systems with asymmetric saturation actuators,'' \emph{IEEE
  transactions on neural networks and learning systems}, vol.~26, no.~7, pp.
  1532--1538, 2014.

\bibitem{Journals:Chaumette2006}
F.~Chaumette and S.~Hutchinson, ``{Visual servo control. Part I: Basic
  approaches},'' \emph{{IEEE} Robot. Autom. Mag.}, vol.~13, no.~4, pp. 82--90,
  2006.

\bibitem{hosoda1994versatile}
K.~Hosoda and M.~Asada, ``Versatile visual servoing without knowledge of true
  jacobian,'' in \emph{Proceedings of IEEE/RSJ International Conference on
  Intelligent Robots and Systems (IROS'94)}, vol.~1.\hskip 1em plus 0.5em minus
  0.4em\relax IEEE, 1994, pp. 186--193.

\end{thebibliography}
\bibliographystyle{IEEEtran}

\end{document}